%% file: main.tex
\documentclass{article}
\usepackage{amssymb}
\usepackage{tikz}
\usepackage{soul}
\usepackage{verbatim}
\usepackage[small]{caption}
\usetikzlibrary{shapes.geometric, arrows.meta, positioning, calc, fit, backgrounds}

\PassOptionsToPackage{sort}{natbib} %
\usepackage[final]{corl_2026} %
\usepackage{latexml} %
\iflatexml
  \usepackage{authblk}
  \newcommand{\keywords}[1]{\par\noindent\textbf{Keywords:} #1}
\fi
\usepackage{xcolor}
\definecolor{customRed}{RGB}{153,27,30}
\hypersetup{
    colorlinks=true,
    citecolor=customRed,
    linkcolor=customRed,
    urlcolor=customRed
}

\input{inc-math}

\input{inc-packages}
\input{inc-macros}

\title{Beyond the End-Effector: Learning Whole-Arm Contact Manipulation from Human Demonstrations}

\iflatexml %
\author[1]{Bo-Ruei Huang}
\author[2]{Daniel Seita}
\author[3]{Erdem B{\i}y{\i}k}
\affil[1]{Thomas Lord Department of Computer Science, University of Southern California \\ \texttt{borueihu@usc.edu}}
\affil[2]{Thomas Lord Department of Computer Science, University of Southern California \\ \texttt{seita@usc.edu}}
\affil[3]{Thomas Lord Department of Computer Science, University of Southern California \\ \texttt{biyik@usc.edu}}
\else
\author{
  Bo-Ruei Huang, Daniel Seita, Erdem B{\i}y{\i}k \\
  Thomas Lord Department of Computer Science, University of Southern California \\
  \texttt{\{borueihu, seita, biyik\}@usc.edu}
}
\fi

\begin{document}

\raggedbottom %

\maketitle

\input{text/0-abstract}
\input{text/1-introduction}
\input{text/2-related_work}

\input{text/3-method}

\input{text/4-experiment}
\input{text/5-conclusion}

\input{text/6-limitations}

\input{text/7-acknowledgments}

\setlength{\bibsep}{5pt}
\bibliography{references}

\clearpage
\input{text/appendix}

\end{document}

%% file: inc-math.tex
\usepackage{amsmath,amsfonts,bm}

\def\eqref#1{equation~\ref{#1}}

\def\1{\bm{1}}

\DeclareMathAlphabet{\mathsfit}{\encodingdefault}{\sfdefault}{m}{sl}
\SetMathAlphabet{\mathsfit}{bold}{\encodingdefault}{\sfdefault}{bx}{n}

\DeclareMathOperator*{\argmax}{arg\,max}

%% file: inc-packages.tex
\usepackage{color,xcolor}
\usepackage{epsfig}
\usepackage{graphicx}

\usepackage[export]{adjustbox}
\usepackage{array}
\usepackage{booktabs}
\usepackage{colortbl}
\usepackage{wrapfig}
\usepackage{hhline}
\usepackage{multirow}
\usepackage{subcaption} %
\usepackage{wrapfig}
\usepackage{floatflt}
\usepackage{siunitx}
\usepackage{caption}

\usepackage{amsmath,amsfonts,amssymb,amsthm}
\usepackage{mathtools}  %
\usepackage{bm}
\usepackage{nicefrac}
\usepackage{microtype}
\usepackage[T1]{fontenc}
\usepackage[ansinew]{inputenc}

\usepackage{changepage}
\usepackage{extramarks}
\usepackage{fancyhdr}
\usepackage{lastpage}
\usepackage{setspace}
\usepackage{soul}
\usepackage{xspace}

\usepackage{url}

\usepackage{enumerate}
\usepackage{todonotes} %
\usepackage{enumitem}  %

\usepackage{titlesec}

\usepackage{makecell}

\usepackage{algorithm}
\usepackage[noend]{algpseudocode}

\usepackage{framed}
\definecolor{shadecolor}{gray}{0.95}

\usepackage{xparse}
\usepackage{etoolbox}
\usepackage{listings}
\usepackage{tcolorbox}
\tcbuselibrary{skins,listings}
\usepackage{pgfplots}
\pgfplotsset{compat=1.18}
\usepgfplotslibrary{groupplots}

\usepackage{float} %

%% file: inc-macros.tex
\DeclareMathAlphabet{\mathbbb}{U}{bbold}{m}{n}

\newcolumntype{L}[1]{>{\raggedright\let\newline\\\arraybackslash\hspace{0pt}}m{#1}}
\newcolumntype{C}[1]{>{\centering\let\newline\\\arraybackslash\hspace{0pt}}m{#1}}
\newcolumntype{R}[1]{>{\raggedleft\let\newline\\\arraybackslash\hspace{0pt}}m{#1}}

\usepackage{pifont}%

\newcommand{\sect}[1]{Section~\ref{#1}}
\newcommand{\sectapp}[1]{Appendix~\ref{#1}}

\newcommand{\eqn}[1]{Equation~(\ref{#1})}
\newcommand{\fig}[1]{Figure~\ref{#1}}
\newcommand{\tbl}[1]{Table~\ref{#1}}
\newcommand{\alg}[1]{Algorithm~\ref{#1}}

\newcommand{\degree}{\ensuremath{^\circ}\xspace}
\newcommand{\ignore}[1]{}

\makeatletter
\DeclareRobustCommand\onedot{\futurelet\@let@token\@onedot}
\def\@onedot{\ifx\@let@token.\else.\null\fi\xspace}

\def\eg{e.g\onedot} 
\def\ie{i.e\onedot}

\makeatother

\definecolor{RayColor}{rgb}{0,0.08,1}
\definecolor{URL}{HTML}{0000EE}
\definecolor{MyDarkGreen}{rgb}{0.02,0.6,0.02}
\definecolor{MyDarkRed}{rgb}{0.8,0.02,0.02}
\definecolor{MyDarkOrange}{rgb}{0.40,0.2,0.02}
\definecolor{MyPurple}{RGB}{111,0,255}
\definecolor{MyRed}{rgb}{1.0,0.0,0.0}
\definecolor{MyGold}{rgb}{0.75,0.6,0.12}
\definecolor{MyDarkgray}{rgb}{0.66, 0.66, 0.66}

\newcommand{\model}{WACL\xspace}
\newcommand{\modelfull}{\textit{Whole-Arm Contact Learning}\xspace}

\newcommand{\taskfigref}{\fig{fig:tasks}}

\newcommand{\myparagraph}[1]{\noindent\textbf{#1}}

\theoremstyle{plain}%

\usepackage{titlesec}
\titlespacing*{\section}{0pt}{3pt plus 2pt minus 2pt}{3pt plus 2pt minus 2pt}
\titlespacing*{\subsection}{0pt}{2pt plus 2pt minus 2pt}{2pt plus 2pt minus 2pt}
\titlespacing*{\subsubsection}{0pt}{1pt plus 1pt minus 1pt}{1pt plus 1pt minus 1pt}

\newcommand{\opencoffee}{\textsc{OpenCoffee}\xspace}
\newcommand{\closecabinet}{\textsc{CloseCabinet}\xspace}
\newcommand{\closefaucet}{\textsc{CloseFaucet}\xspace}
\newcommand{\pressbutton}{\textsc{PressButton}\xspace}

\newcommand{\vspacesection}[1]{\section{#1}}
\newcommand{\vspacesubsection}[1]{\subsection{#1}}
\newcommand{\vspacesubsubsection}[1]{\subsubsection{\textbf{#1}}}

\definecolor{promptshade}{RGB}{246,248,251}
\newsavebox{\promptboxsave}
  {\vspace{2pt}\end{minipage}%
  \end{lrbox}%
  \colorbox{promptshade}{\usebox{\promptboxsave}}\par\medskip}

\definecolor{promptbar}{RGB}{44,62,92}
\definecolor{prompthead}{RGB}{31,78,121}
\definecolor{promptvar}{RGB}{176,48,40}
\lstdefinestyle{prompt}{
  basicstyle=\fontfamily{lmtt}\fontsize{7}{7.6}\selectfont,
  columns=fullflexible, keepspaces=true, upquote=true, showstringspaces=false,
  breaklines=true, breakatwhitespace=true, breakautoindent=true, breakindent=0pt,
  prebreak={}, postbreak={}, aboveskip=0pt, belowskip=0pt,
  moredelim=**[is][\color{promptvar}]{@}{@},
  moredelim=[is][\sffamily\bfseries\color{prompthead}]{|}{|},
  literate={->}{{$\rightarrow$}}2 {--}{{\textemdash}}1 {~}{{\textbullet}}1,
}
\iflatexml %
\lstnewenvironment{promptlisting}[1]{\lstset{style=prompt, frame=single, title={\textbf{\def\hfill{ --- }#1}}}}{}
\else
\newtcblisting{promptlisting}[1]{
  enhanced, listing only, listing options={style=prompt}, width=\linewidth,
  arc=2pt, boxrule=0.6pt, colback=promptshade, colframe=promptbar, colbacktitle=promptbar,
  coltitle=white, fonttitle=\sffamily\bfseries\small, title={#1},
  left=5pt, right=5pt, top=4pt, bottom=4pt, toptitle=2pt, bottomtitle=2pt}
\fi

%% file: text/0-abstract.tex
\begin{abstract}
Robot manipulation has traditionally relied on end-effector-centric interactions.
In contrast, humans routinely exploit whole-arm contacts to improve efficiency, robustness, and adaptability, especially when the hands are occupied or when precise grasping is unnecessary.
Enabling similar capabilities in robots remains challenging: the robot must select where on its arm to make contact, generate effective non-prehensile interactions, and satisfy task-specific pose or safety constraints on the end-effector.
We propose \modelfull (\model), a hierarchical framework that describes each whole-arm contact task with three \emph{robot-agnostic} representations and hands them to a quasi-static surface-contact optimizer: semantic keypoints that summarize the scene, a reference contact trajectory that specifies where and how the arm presses on the object, and safety-aware path constraints that a vision-language model generates from the task description to protect the end-effector's task.
The robot's kinematic chain enters only in this optimizer, which grounds these representations onto the arm by jointly searching over which link makes contact, where on that link, and the joint configuration that realizes it.
This decomposition lets \model learn the contact trajectory from $10$ \emph{bare-hand human} demonstrations per task collected without teleoperation, while new end-effector constraints change only the optimizer's input rather than any trained component.
Real-world experiments on a 7-DoF arm across four whole-arm tasks show that \model sustains contact under end-effector constraints where a strong link-pose baseline fails ($78\%$ vs.\ $33\%$ success) and generalizes to new constraints without retraining.
Website: \href{https://wacl-robot.github.io}{https://wacl-robot.github.io}.
\end{abstract}
\keywords{Robot manipulation, Combination of learning and planning in robotics, Whole-arm manipulation, Learning from human demonstrations}

%% file: text/1-introduction.tex
\vspacesection{Introduction}

In our daily lives, manipulation is rarely confined to the fingertips.
When our hands are already occupied, such as when carrying a load or holding a cup of coffee, we routinely recruit the rest of the arm to act on the environment, nudging a door open with an elbow or steadying a box against a forearm.
This ``Whole-Arm Manipulation'' (WAM)~\citep{eberman1989whole} lets us take on a second task without abandoning the first: the hand keeps the cup upright---the \emph{primary} task---while the arm presses a button or pushes a drawer shut---the \emph{secondary} task.
This multi-task setting introduces a constraint: whatever the arm does, it must not disturb what the hand is already committed to.

Robot manipulation, however, remains overwhelmingly end-effector-centric.
Pre-trained vision-language-action models (VLAs) typically output gripper-pose actions~\citep{zitkovich2023rt, kim24openvla, intelligence2025pi0, intelligence2025pi05, bjorck2025gr00t, geminirobotics2025, lee2025molmoact} and treat the arm as a mechanism for transport, so arm-environment contact is usually avoided rather than exploited. %
This discards useful kinematic redundancy and limits the robot in constrained or cluttered settings, where whole-arm contact may be necessary because a gripper-only motion is infeasible.
Learning whole-arm manipulation end-to-end is also difficult for two compounding reasons.
First, kinematically similar contacts on different links (\eg, elbow, forearm, upper arm) produce substantially different pixel observations and joint configurations, so transferring contact behavior across links is difficult for a policy trained on raw inputs, a within-robot generalization gap.
Second, while people regularly engage in whole-arm contact, teleoperating a 7-DoF arm into the same contacts is slow and unintuitive, and typically requires a leader-follower system~\citep{wu2024gello} or an exoskeleton~\citep{fang2024airexo}.
Bare-hand human demonstrations are therefore a more natural data source, but they provide no robot actions~\citep{lepert2025phantomtrainingrobots,kareer2025egomimic,huang2024difo,hong2026hand,sontakke2023roboclip} and introduce a human-to-robot embodiment gap.

We propose \modelfull (\model), a hierarchical framework that closes both gaps by describing each task with three robot-agnostic representations and handing them to a surface-contact trajectory optimizer: \emph{semantic keypoints} that summarize the scene, a \emph{reference contact trajectory} that specifies where and how the arm should press on the object, and \emph{safety-aware path constraints} on the end-effector that keep its own task valid.
The robot's kinematic chain enters only once, in this optimizer, which grounds these representations onto the arm by jointly searching over which link makes contact, where on that link, and the joint configuration that realizes it, as illustrated in \fig{fig:pipeline}.
This decomposition lets \model learn contact trajectories from bare-hand human demonstrations, while new end-effector constraints change only the optimizer's input, not the trained policy.

We evaluate \model on four real-world whole-arm tasks on a 7-DoF xArm, each requiring a secondary contact action while preserving a primary end-effector task (\eg, turning a faucet off while keeping a wet pad above the sink).
Without end-effector constraints, \model performs on par with or above a strong keypoint-conditioned baseline, KALM~\citep{fang2025kalm} adapted to regress the contact link's pose; under constraints, \model more than doubles the baseline's success rate ($78\%$ vs.\ $33\%$) by re-routing the contact to a link and anchor that keep the primary task intact.
\model further handles a second, simultaneous constraint with no retraining and transfers zero-shot to a second arm, a Franka FR3, with only the optimizer's robot model changed; an ablation shows that the sliding surface anchor accounts for part of its constrained performance.

Our contributions are summarized as follows:
\begin{itemize}[leftmargin=*,nosep,topsep=-3pt]
\item \model, a hierarchical framework that decomposes whole-arm manipulation into three robot-agnostic representations: semantic keypoints, a reference contact trajectory, and safety-aware path constraints, so that the robot's kinematic chain is grounded in a single downstream optimizer.
\item A robot-agnostic data pipeline that learns the contact trajectory from bare-hand, multi-view human demonstrations, with no whole-arm teleoperation.
\item A surface-contact trajectory optimizer that jointly searches over the contact link, a sliding surface anchor, and the joint configuration under path constraints generated by a vision-language model (VLM), %
with real-world experiments showing sustained contact under end-effector constraints.
\end{itemize}

%% file: text/2-related_work.tex
\vspacesection{Related work}
\label{sec:related_work}

\myparagraph{Whole-body manipulation and sensing.}
Whole-body control for humanoids, quadrupeds, and mobile manipulators coordinates many degrees of freedom~\citep{ha2024umilegs,fu2023deep,jeon2023learning,cheng2024express}, but its focus is balance, locomotion, and floating-base coordination rather than exploiting limb surfaces for functional interaction.
A separate line of work in contact-rich manipulation explores tactile feedback~\citep{wistreich2025dexskin} and compliant skins~\citep{luo2021learning,xu2024cushsensesoftstretchable,si2023robotsweater} to enable richer body-environment interaction, but these assume specialized hardware.
A smaller body of work treats the arm's surface itself (\eg, forearm, elbow) as the manipulation interface.
Learning-based methods collect joint-space demonstrations through an exoskeleton kinematically matched to the robot's arm~\citep{fang2024airexo,fang2025airexo2} or hand~\citep{fang2026dexop}, or train per-task sim-to-real reinforcement-learning policies for a compliant, tactile-skinned humanoid~\citep{barreiros2025learning}.
Model-based methods plan contacts from known object models by global search over smoothed quasi-dynamic contact models~\citep{pang2023global}, tensor-factorized tree search~\citep{xue2025monte}, or continuous optimization over an explicit robot-surface representation~\citep{leve2024explicit}.
To our knowledge, none holds a concurrent end-effector task while the arm makes contact.
Our framework instead learns the contact trajectory from bare-hand human demonstrations, needs no robot demonstrations, object model, or tactile sensing, and enforces the end-effector task inside the optimizer that chooses the contact link.

\myparagraph{Vision-language models (VLMs) for robotics.}
Foundation models, including VLMs, have proven useful for robotic reasoning, enabling systems to perform spatial planning~\citep{huang2023voxposer, fangandliu2024moka}, synthesize control code~\citep{codeaspolicies2022}, craft reward functions~\citep{ma2023eureka,wang2024rlvlmf}, and specify geometric constraints~\citep{huang2024rekep,huang2024copa,chen2024autotamp}. 
Recent work also moves beyond collision-free reasoning to use VLMs for contact semantics, inferring which parts of an environment can tolerate interaction~\citep{ling2026impact}.
These methods constrain or plan the end-effector; none selects a contact link on the arm or protects a task the end-effector is already performing.
Hierarchical frameworks ground high-level semantic plans in learned low-level policies~\citep{saycan2022arxiv, driess2023palm, fang2025kalm, zhang2026peek}.
We apply VLM-generated constraints at the \emph{link level}: they must hold on one link while a different link makes contact, and changing them never retrains the policy.

%% file: text/3-method.tex
\section{Method}
\label{sec:method}

\input{figs/framework} %

We propose \modelfull (\model), a hierarchical framework that decomposes whole-arm manipulation into three stages. %
\sect{sec:formulation} defines the task representations and the constrained optimization they induce.
\model learns the reference contact trajectory from bare-hand demonstrations (\sect{sec:traj_gen}), instantiates path constraints from the task description and scene observation (\sect{sec:constraint_gen}), and optimizes a feasible whole-arm trajectory (\sect{sec:traj_opt}).  
\sect{sec:inference} combines the three stages into the end-to-end inference pipeline.
See \fig{fig:pipeline} for an overview. %

\vspacesubsection{Problem formulation}
\label{sec:formulation}
\label{sec:abstraction}

\model represents each task with three \emph{robot-agnostic} representations, deliberately free of any particular robot's kinematics.
The \emph{semantic keypoints} $K_\text{obs} = \{\mathbf{p}_i\}_{i=1}^{n_\text{kp}} \in \mathbb{R}^{n_\text{kp} \times 3}$ summarize the observation as 3D scene points that localize task-relevant geometry independent of the robot's pose (\sect{sec:traj_gen}).
The \emph{reference contact trajectory} $\mathcal{T}^{*} = \{(\mathbf{p}^{*}_t, \mathbf{n}^{*}_t)\}_{t=1}^{H}$ specifies a world-frame contact position and surface normal at each of $H$ waypoints, with no commitment to which link makes contact or to any joint configuration (\sect{sec:traj_gen}).
The \emph{safety-aware path constraints} $\{\mathcal{C}_k\}$ are typed requirements on link-level quantities (\eg, ``the gripper axis stays within $25\degree$ of world up''), evaluated at every waypoint (\sect{sec:constraint_gen}).

The optimizer acts on a robot with configuration space $\mathcal{Q} \subset \mathbb{R}^{n_q}$ and joint configuration $q_t \in \mathcal{Q}$ over a horizon of $H$ waypoints covering the contact segment.
To let it decide \emph{which} link makes contact, we expose the contact-capable links explicitly: each link $\ell \in \mathcal{L}$ carries a collision mesh $\mathcal{M}_\ell$, and their disjoint union $\mathcal{M} = \bigsqcup_{\ell \in \mathcal{L}} \mathcal{M}_\ell$ is the robot's contact surface, with every point tagged by the link it belongs to.
A \emph{surface anchor} is a point $\mathbf{u} \in \mathcal{M}_\ell$ in $\ell$'s local frame, whose world-frame position and normal at any $q$ follow from forward kinematics (\sect{sec:traj_opt}).

\myparagraph{Objective.}
At inference the system receives an observation $O = (I, d_\text{task})$, a calibrated RGB-D image and a natural-language task description, from which it produces $(K_\text{obs}, \mathcal{T}^{*}, \{\mathcal{C}_k\})$.
Given $\mathcal{T}^{*}$ and $\{\mathcal{C}_k\}$, we seek a joint trajectory $q_{1:H}$, a contact link $\ell \in \mathcal{L}$, and a per-waypoint anchor sequence $\mathbf{u}_{1:H}$ such that the world-frame contact at each waypoint tracks $(\mathbf{p}^{*}_t, \mathbf{n}^{*}_t)$, every $\mathcal{C}_k$ is satisfied, and the trajectory is kinematically and collision feasible.
The free-space approach from the robot's home pose to the start of the contact segment is planned separately (\sect{sec:traj_opt}).

\vspacesubsection{Contact trajectory learning}
\label{sec:traj_gen}

Because $\mathcal{T}^{*}$ names only a contact position and normal, a bare hand under multi-camera RGB-D capture can produce it directly.
This subsection describes the offline pipeline that turns such demonstrations into the $(K_\text{obs}, \mathcal{T}^{*})$ pairs used to train the keypoint-conditioned policy $\pi_\theta$.

\vspacesubsubsection{Offline data pipeline}
\label{sec:traj_gen_pipeline}

\myparagraph{Multi-view hand demonstration capture.}
A human performs each task bare-hand under $C$ calibrated RGB-D cameras.
Each demonstration is automatically segmented to its contact phase by a fingertip-velocity threshold; the operator can mark the contact boundary by pausing without extra cost.
The human contacts the object with a fingertip, a point proxy for the robot's link-surface contact.

\myparagraph{Per-view hand pose lifting.}
For each per-camera RGB-D stream, we segment the hand with SAM3~\citep{carion2025sam} and regress MANO~\citep{MANO:SIGGRAPHASIA:2017} parameters using the 4D hand tracker HaPTIC~\citep{ye2025predicting}.
The tracks are temporally coherent, but each view is monocular and therefore inaccurate in depth~\citep{lepert2025phantomtrainingrobots}, while the contact trajectory needs accurate 3D fingertip positions.

\myparagraph{Multi-view hand fusion.}
To address the problem, we fuse the per-view estimates by triangulating the contact fingertip vertex across the $C$ calibrated views via per-point weighted-midpoint ray triangulation~\citep{hartley1997triangulation}, with each camera weighted by the silhouette intersection-over-union (IoU) between the reprojected MANO mesh and the SAM3 hand mask in that view.
Because each camera's depth uncertainty lies along its viewing ray, triangulation drives the residual into directions the other cameras constrain laterally, yielding a single world-frame fingertip trajectory consistent across views.

\myparagraph{Reference contact trajectory extraction.}
We resample the fused fingertip trajectory to the fixed horizon $H$ via cubic spline, and pair each sampled position $\mathbf{p}^{*}_t$ with the corresponding MANO surface normal $\mathbf{n}^{*}_t$ at the contact fingertip to obtain the reference contact trajectory $\mathcal{T}^{*} = \{(\mathbf{p}^{*}_t, \mathbf{n}^{*}_t)\}_{t=1}^{H}$ for the demonstration.
The resulting $\mathcal{T}^{*}$ does not refer to the robot, so the same demonstrations serve any arm; \sectapp{app:data} details capture, fusion, and extraction.

\vspacesubsubsection{Keypoint distillation and policy training}
\label{sec:traj_gen_policy}

\myparagraph{Semantic keypoint distillation.}
The observation representation $K_\text{obs}$ is produced by the KALM keypoint distillation pipeline~\citep{fang2025kalm}.
One demonstration is selected as the reference and its task-relevant region is segmented by a VLM-driven SAM3 prompt; in every other demonstration, DINOv2~\citep{oquab2023dinov2} features and FPFH~\citep{rusu2009fast} geometric consistency localize a fixed set of $n_\text{kp}$ keypoints corresponding to the reference.
At inference, the same matcher $\Phi_\text{KP}$ is applied to a live RGB-D view and recovers each keypoint by maximizing feature similarity against the reference,
\begin{equation}
\mathbf{p}_i = \argmax_{\mathbf{p}} \,\langle F_\text{obs}(\mathbf{p}),\, F_\text{ref}(\mathbf{k}_i) \rangle, \qquad K_\text{obs} = \{\mathbf{p}_i\}_{i=1}^{n_\text{kp}},
\end{equation}
where $F_\text{obs}$ and $F_\text{ref}$ are the combined DINOv2 and FPFH features of the query and reference views and $\{\mathbf{k}_i\}_{i=1}^{n_\text{kp}}$ are the reference keypoints.

\myparagraph{Keypoint-conditioned policy.}
We adopt KALM~\citep{fang2025kalm} as a conditional trajectory diffusion policy $\pi_\theta$ that models $p(\mathcal{T}^{*} \mid K_\text{obs})$, trained with a noise-prediction objective in the style of Diffuser~\citep{janner2022planning} on the $(K_\text{obs}, \mathcal{T}^{*})$ pairs from \sect{sec:traj_gen_pipeline}.
At inference, the policy iteratively denoises Gaussian noise into a reference contact trajectory, $\mathcal{T}^{*} \sim \pi_\theta(\cdot \mid K_\text{obs})$.

\vspacesubsection{Safety-aware path constraint generation}
\label{sec:constraint_gen}

The reference contact trajectory $\mathcal{T}^{*}$ specifies \emph{where} on the scene a chosen link must make contact, but says nothing about the rest of the arm.
Whole-arm manipulation typically arises when the end-effector is already committed to a primary task, such as carrying a cup, and the contact action must not disturb it.
The path constraints $\{\mathcal{C}_k\}$ encode this remaining requirement: they restrict the pose of links \emph{other than} the contact link throughout the contact segment so that the primary end-effector task is preserved regardless of what the contact link is doing.
We populate $\{\mathcal{C}_k\}$ at inference from a fixed, typed constraint vocabulary whose parameters are filled from an open-vocabulary scene state via two GPT-5.5 calls, one for scene state estimation and one for constraint instantiation.

\myparagraph{Constraint primitives.}
Each constraint $\mathcal{C}_k$ instantiates a primitive from a small fixed vocabulary, a requirement on one link's axis or position, with parameters $\boldsymbol{\phi}_k$.
Its violation $c_k(q; \boldsymbol{\phi}_k) \ge 0$ (written $c_k(q)$ below) is zero when the requirement holds at $q$ and grows with the amount by which it is missed.
The optimizer uses this one quantity twice: added to the objective as a soft cost, so partial violations have a gradient to descend, and as the hard test $c_k(q_t) \le \tau_k$ at every waypoint, so no weighting of the contact objective can push a violation past the tolerance $\tau_k$.
Primitive types per task are listed in \sect{sec:exp_setup}; deployed parameters and tolerances in \sectapp{app:impl_constraint}.

\myparagraph{Open-vocabulary scene state estimation.}
The first VLM call takes $(I, d_\text{task})$ and enumerates the task-relevant objects in the scene via chain-of-thought decomposition~\citep{wei2022chain}.
Each named object is segmented by SAM3~\citep{carion2025sam} into a 2D mask, back-projected into a 3D point cloud using the depth channel of $I$, and reduced to an oriented bounding box.
The result is a scene state $\mathcal{S} = \{(\text{label}_j, \mathcal{B}_j)\}$ of named entities whose bounding boxes fill the primitives' geometric parameters.

\myparagraph{Constraint instantiation.}
The second VLM call receives $d_\text{task}$, the scene state $\mathcal{S}$, and the world-frame end-effector pose from forward kinematics.
It selects a primitive for each requirement in $d_\text{task}$, fills $\boldsymbol{\phi}_k$ from $\mathcal{S}$, and returns a typed list parsed into $\{\mathcal{C}_k\}$; \sectapp{app:constraint} gives the prompts.

\vspacesubsection{Surface-contact trajectory optimization}
\label{sec:traj_opt}

This subsection instantiates the constrained optimization of \sect{sec:formulation}: jointly determine a contact link $\ell$, an anchor sequence $\mathbf{u}_{1:H}$, and a joint trajectory $q_{1:H}$ so that the resulting world-frame contact tracks $\mathcal{T}^{*}$ at every waypoint while satisfying every $\mathcal{C}_k$.
The three decision variables are qualitatively different --- $\ell$ is discrete, each $\mathbf{u}$ is confined to a 2D mesh surface, and each $q$ is continuous --- so a gradient-only formulation is poorly suited to the joint search.
We instead adopt a sample-and-search formulation: discretize the per-link contact mesh into a candidate anchor set, solve a surface-contact inverse kinematics problem at each anchor and waypoint to obtain admissible joint configurations, and assemble these into a trajectory via beam search.
Our first attempt, a continuous program over the joint trajectory and a contact point blended across mesh triangles, rarely converged, as \sectapp{app:impl_opt} details.
Sampling anchors and solving per-waypoint inverse kinematics (IK) in parallel on the GPU prunes infeasible branches early, at the cost of a greedy search smoothed only by the step cost of \eqn{eq:step_cost}.
For link $\ell$ at joint configuration $q$, forward kinematics returns the $SE(3)$ transform
\begin{equation}
T_\ell(q) = \begin{bmatrix} R_\ell(q) & t_\ell(q) \\ \mathbf{0}^\top & 1 \end{bmatrix}, \qquad \mathbf{p}(q, \mathbf{u}) = T_\ell(q)\,\mathbf{u}, \qquad \mathbf{n}(q, \mathbf{u}) = R_\ell(q)\,\mathbf{n}^\text{local}(\mathbf{u}),
\label{eq:fk}
\end{equation}
where $\mathbf{n}^\text{local}(\mathbf{u})$ is the outward mesh normal at the link-local anchor $\mathbf{u}$, and the full objective is
\begin{equation}
\min_{q_{1:H},\,\ell,\,\mathbf{u}_{1:H}} \sum_{t=1}^H J_\text{contact}\big(q_t,\mathbf{u}_t;\,\mathbf{p}^{*}_t,\mathbf{n}^{*}_t\big) \quad \text{s.t.}\quad c_k(q_t)\le\tau_k\ \forall k,t;\ \ q_t\in\mathcal{Q};\ \ \text{collision-free},
\end{equation}
with $J_\text{contact}$ the per-waypoint surface-contact cost defined below.

\input{figs/tasks} %
\myparagraph{Reachable anchor set per link.}
For each contact link $\ell$, we precompute a candidate anchor set on $\mathcal{M}_\ell$ by farthest-point sampling~\citep{eldar1997farthest} its collision mesh, the same geometry the planner uses for self-collision checks so that every candidate is physically meaningful.
We then prune anchors that are unreachable in practice: those obstructed by adjacent links along their outward normal, those lying inside the signed-distance volume of a neighboring link, and those recessed inside the convex hull of $\mathcal{M}_\ell$. The remaining anchors form the discrete search space the optimizer expands over.

\myparagraph{Surface-contact inverse kinematics.}
For each (waypoint $t$, candidate anchor $\mathbf{u}$) pair, we solve a surface-contact IK problem that drives the chosen link to track $(\mathbf{p}^{*}_t, \mathbf{n}^{*}_t)$ under the cost
\begin{equation}
J_\text{contact}(q,\mathbf{u};\,\mathbf{p}^{*},\mathbf{n}^{*}) = w_\text{pos}\,\|\mathbf{p}(q,\mathbf{u}) - \mathbf{p}^{*}\|^2 + w_\text{align}\,\big[\cos\theta_\text{dead} - \langle \mathbf{n}(q,\mathbf{u}),\,\mathbf{n}^{*}\rangle\big]_+,
\label{eq:contact_cost}
\end{equation}
combining a quadratic position residual with a hinge alignment cost on the surface normal.
The hinge zeros the alignment penalty whenever the angular error is within the deadzone $\theta_\text{dead}$, so IK can trade a small tilt of the normal for position accuracy.
Standard joint-limit, self-collision, scene-collision, and $\{c_k\}$ soft costs are added to the IK objective~\citep{sundaralingam2023curobo}; solutions that exceed a tolerance $\tau_k$ or collide with the scene are discarded, leaving admissible $(t, \mathbf{u}, q)$ triples for beam expansion.

\myparagraph{Beam search and trajectory assembly.}
Per-waypoint IK yields admissible $(\mathbf{u}, q)$ candidates independently at each $t$, but an executable trajectory needs adjacent waypoints to agree on joint configuration and to stay close on the contact surface.
Exhaustive enumeration over $H$-step sequences is intractable, so we assemble the trajectory with beam search of width $B$, configured as in \sectapp{app:impl_opt}.
Each partial sequence $(\mathbf{u}_{1:t}, q_{1:t})$ is extended by each admissible anchor at $t{+}1$, scored by the step cost
\begin{equation}
J_\text{step}(q_t, \mathbf{u}_t \!\to\! q_{t+1}, \mathbf{u}_{t+1}) = w_{\Delta q}\,\|q_{t+1} - q_t\| + w_\text{slide}\,d_\text{slide} + w_\text{manip}/\mu(q_{t+1}),
\label{eq:step_cost}
\end{equation}
where $d_\text{slide} = \|\mathbf{u}_{t+1} - \mathbf{u}_t\|$ is the link-local displacement of the surface anchor between $t$ and $t+1$ and $\mu$ is the Yoshikawa manipulability~\citep{yoshikawa1985manipulability}; we keep the top-$B$ extensions until $t = H$.
The search runs in parallel for each candidate link $\ell \in \mathcal{L}$ and the results are pooled, so the optimizer selects the link whose trajectory scores best for the given $\mathcal{T}^{*}$ and $\{\mathcal{C}_k\}$ rather than committing to one in advance.
For each pooled candidate we sample a pre-contact pose by backing off along the contact normal at $t=1$ within a small offset cone, then plan a free-space approach to it from the home pose under $\{\mathcal{C}_k\}$.
We rank the resulting (link, contact, approach) triples by feasibility margin, breaking ties by accumulated slide $\sum_t d_\text{slide}$ and then by joint path length, and execute the top one.

\vspacesubsection{Inference and execution}
\label{sec:inference}

At inference, the stages of \fig{fig:pipeline} run in sequence on a single observation $O = (I, d_\text{task})$; \alg{alg:inference} gives the pseudocode.
The keypoint matcher takes $K_\text{obs}$ from the highest-scoring camera view, the policy samples $\mathcal{T}^{*} \sim \pi_\theta(\cdot \mid K_\text{obs})$, and the constraint pipeline produces $\{\mathcal{C}_k\}$ from the same observation.
The surface-contact optimizer (\sect{sec:traj_opt}) then returns the top-ranked joint trajectory with its free-space approach, which are concatenated and streamed to the robot.

%% file: figs/framework.tex
\definecolor{stage1}{HTML}{F27200}
\definecolor{stage2}{HTML}{070782}
\definecolor{stage3}{HTML}{017100}

\definecolor{capContact}{RGB}{234,155,86}    %
\definecolor{capPredict}{RGB}{90,128,184}    %
\definecolor{capConstraint}{RGB}{161,186,102}%
\definecolor{capOptim}{RGB}{150,123,201}     %

\begin{figure*}[t]
    \centering
    \includegraphics[width=0.9\linewidth,trim={150 50 150 50}]{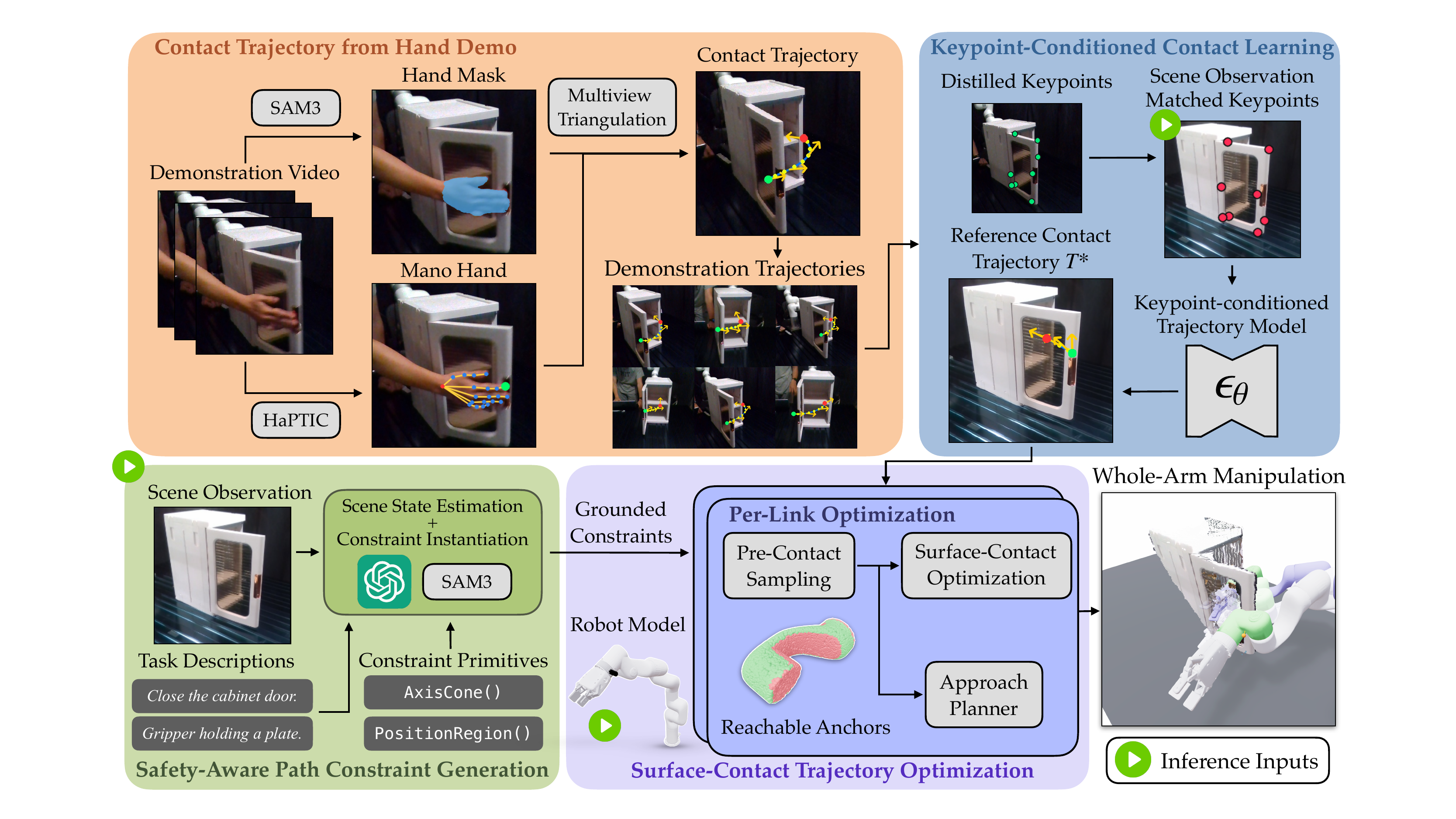}
    \caption{\textbf{\model overview.}
    \model produces three robot-agnostic representations and grounds them onto the robot in a single downstream optimizer.
    \textcolor{capContact}{\textbf{Contact trajectory from hand demonstrations.}}
    A segmentation model and 4D hand tracker fuse bare-hand multi-view RGB-D demonstrations into a \emph{reference contact trajectory} of contact positions and surface normals.
    \textcolor{capPredict}{\textbf{Keypoint-conditioned contact learning.}}
    Semantic keypoints distilled by KALM are matched to the scene, conditioning a trajectory model that generates the contact trajectory.
    \textcolor{capConstraint}{\textbf{Safety-aware path constraint generation.}}
    A VLM grounds a fixed vocabulary of primitives into path constraints that preserve the primary end-effector task.
    \textcolor{capOptim}{\textbf{Surface-contact trajectory optimization.}}
    The optimizer searches each link's reachable anchors with surface-contact inverse kinematics and beam search under the constraints.
    }
    \label{fig:pipeline}
    \vspace{-1.5em}
\end{figure*}

%% file: figs/tasks.tex
\begin{figure*}[t]
    \centering
    \begin{subfigure}[b]{0.245\linewidth}
        \centering
        \includegraphics[width=\linewidth,height=\linewidth,keepaspectratio=false]{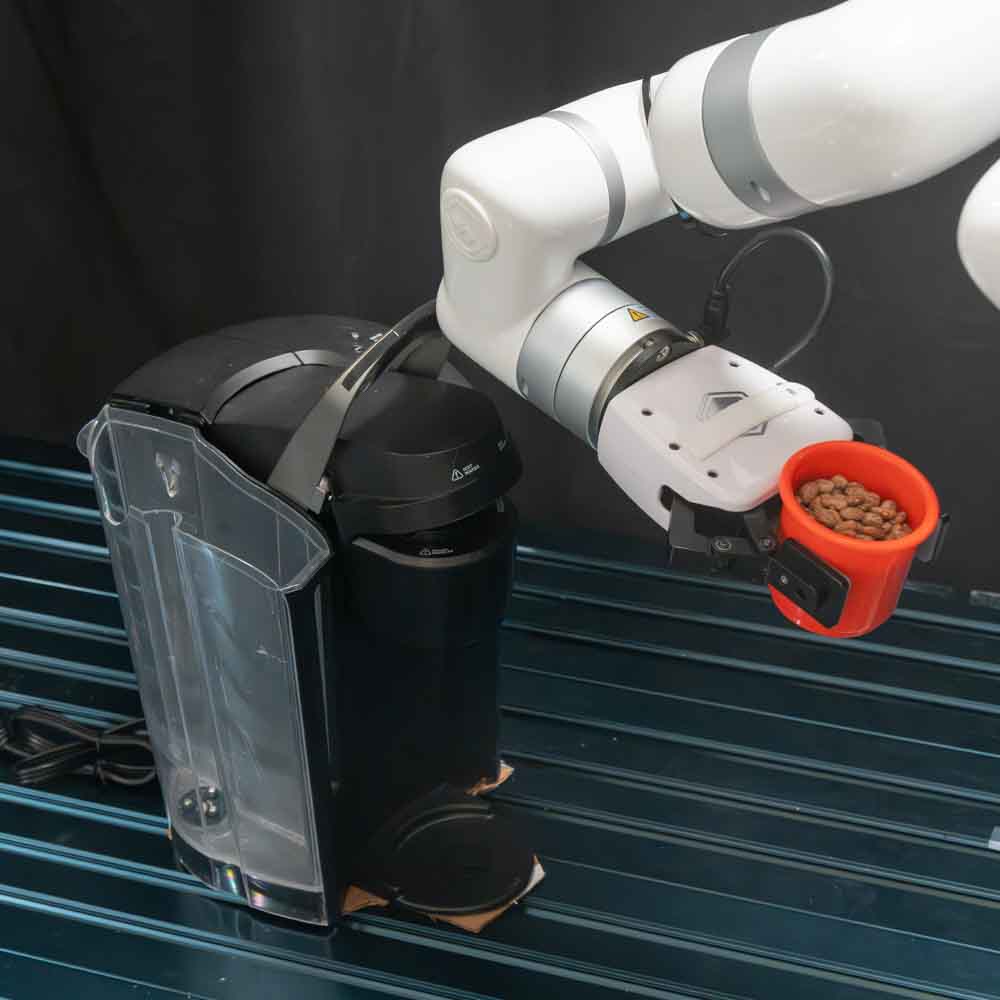}
        \caption{\opencoffee}
        \label{fig:task_coffee}
    \end{subfigure}
    \hfill
    \begin{subfigure}[b]{0.245\linewidth}
        \centering
        \includegraphics[width=\linewidth,height=\linewidth,keepaspectratio=false]{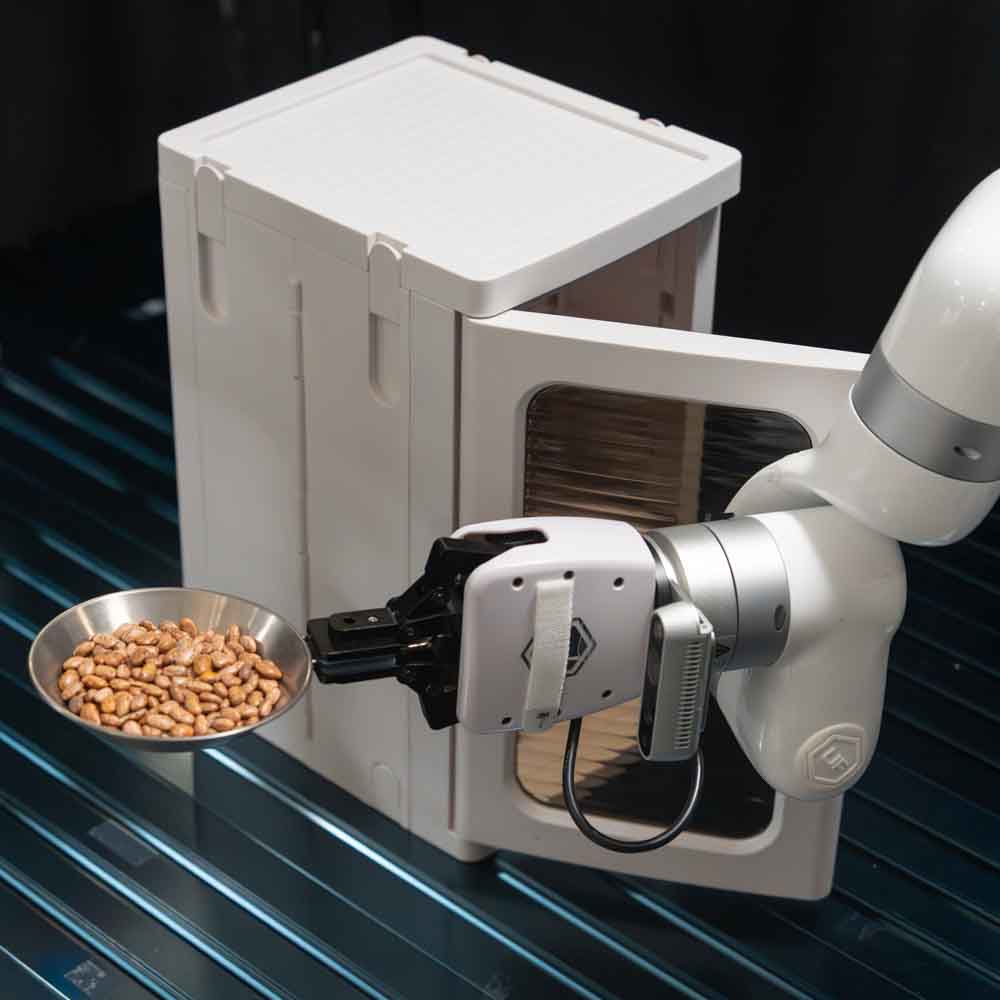}
        \caption{\closecabinet}
        \label{fig:task_cabinet}
    \end{subfigure}
    \hfill
    \begin{subfigure}[b]{0.245\linewidth}
        \centering
        \includegraphics[width=\linewidth,height=\linewidth,keepaspectratio=false]{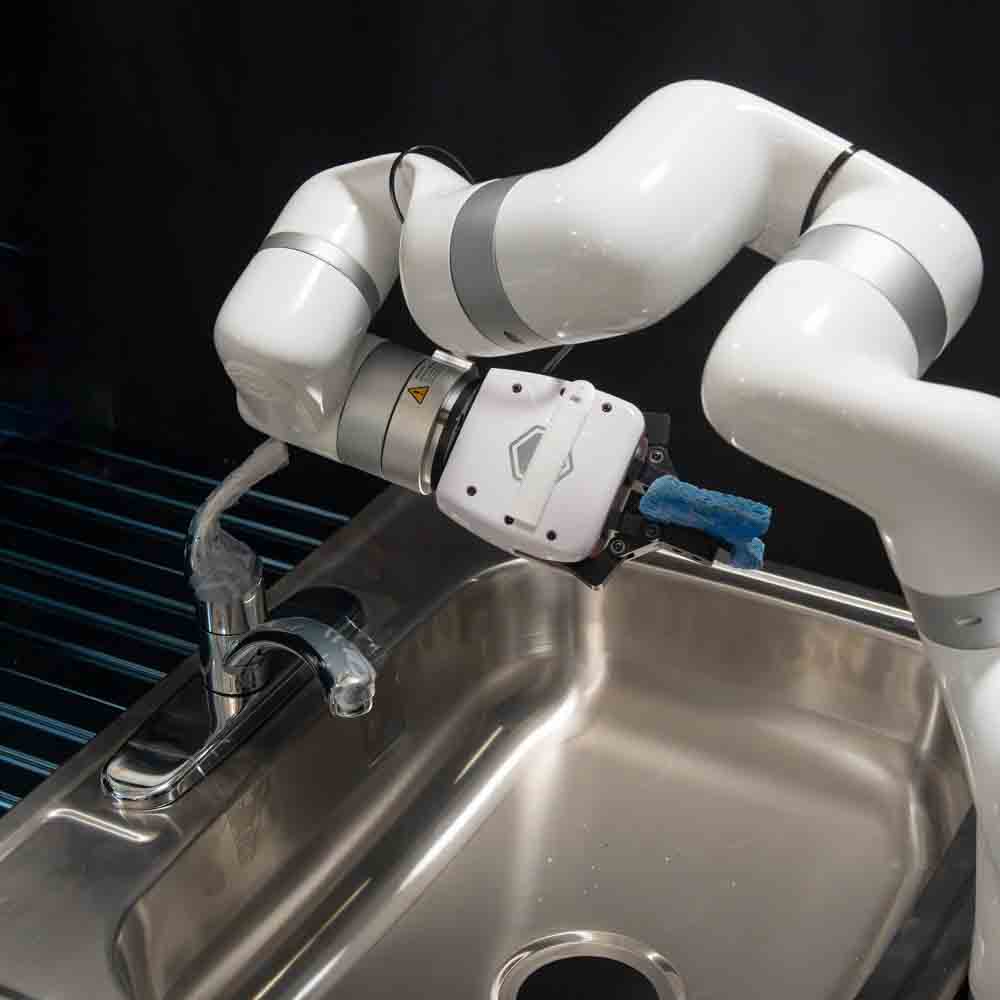}
        \caption{\closefaucet}
        \label{fig:task_faucet}
    \end{subfigure}
    \hfill
    \begin{subfigure}[b]{0.245\linewidth}
        \centering
        \includegraphics[width=\linewidth,height=\linewidth,keepaspectratio=false]{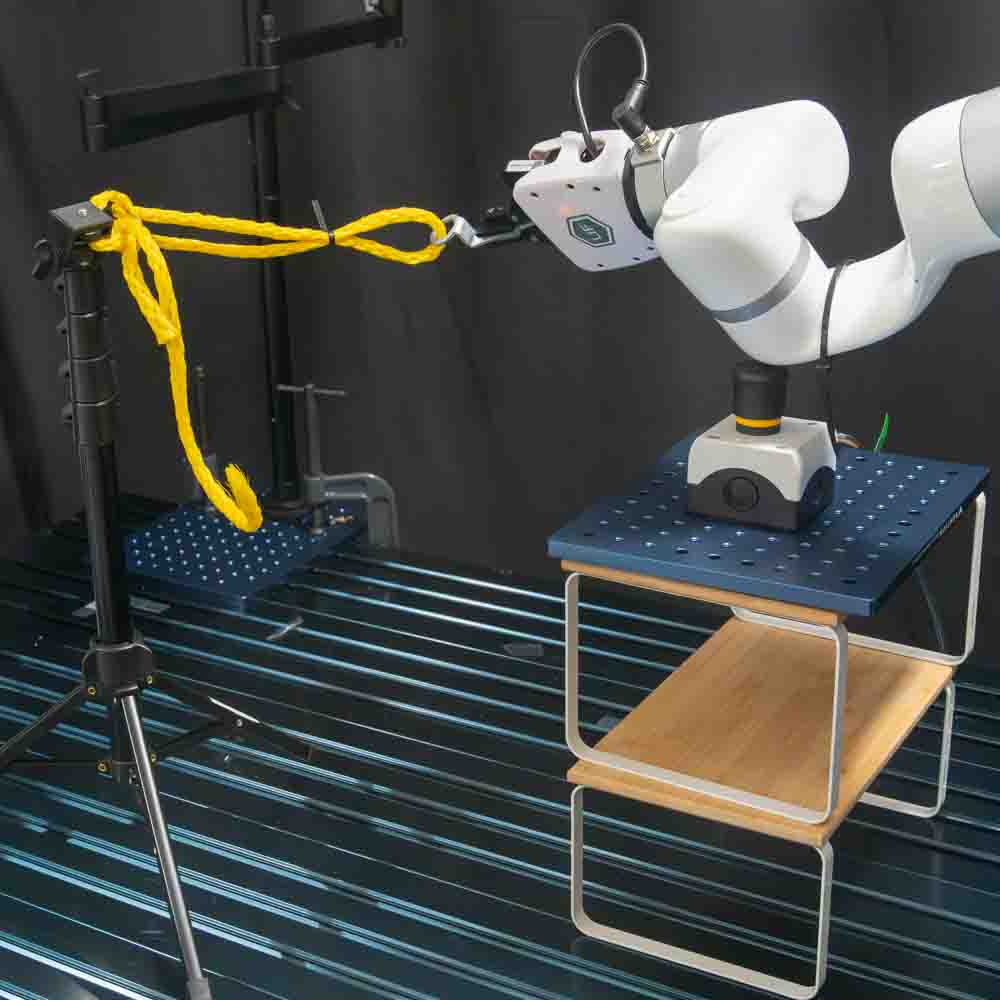}
        \caption{\pressbutton}
        \label{fig:task_button}
    \end{subfigure}
    \caption{\textbf{Evaluation tasks.} Each task pairs a secondary whole-arm contact action with a primary end-effector task that must hold throughout: cup upright, plate level, gripper above the sink, gripper tethered by a rope.}
    \label{fig:tasks}
    \vspace{-1.5em}
\end{figure*}

%% file: text/4-experiment.tex
\vspacesection{Experiments}
\label{sec:experiment}

We design our experiments to answer three questions.
First, does \model outperform a strong link-pose baseline on real-world whole-arm tasks, and does the gap widen under end-effector constraints (\sect{sec:exp_main})?
Second, does it generalize to additional constraints without retraining, and to a different robot (\sect{sec:exp_multi})?
Third, which design choices are responsible for the gain (\sect{sec:exp_ablation})?

\vspacesubsection{Experimental setup}
\label{sec:exp_setup}
\input{tabs/1-main}
\myparagraph{Hardware and workspace.}
All experiments run on a 7-DoF UFactory xArm 7 observed by two calibrated Intel RealSense D435 RGB-D cameras, with planning on one NVIDIA RTX PRO 6000 GPU, described in \sectapp{app:impl_setup}.
The optimizer searches in parallel over the three distal contact-capable links of the xArm, links 4--6, whose collision meshes form the contact surface $\mathcal{M}$ (\fig{fig:contact_links}).

\myparagraph{Tasks and end-effector constraints.}
We evaluate four whole-arm tasks, each constructed so that a primary end-effector task must hold throughout a secondary contact action.
We instantiate two primitives from \sect{sec:constraint_gen}: an \textsc{AxisCone} keeps a link axis within a tolerance angle of a target direction, and a \textsc{PositionRegion} keeps a link position inside a 3D region, here a sphere or a box; the four tasks are (\taskfigref):
\begin{itemize}[leftmargin=*,nosep,topsep=-3pt]
\item \opencoffee: lift the handle of a coffee machine while the gripper holds a cup of water (\textsc{AxisCone}: cup axis aligned with world up).
\item \closecabinet: push a cabinet door closed while the gripper holds a plate (\textsc{AxisCone}: plate normal aligned with world up).
\item \closefaucet: turn a faucet handle off while the gripper holds a wet scouring pad (\textsc{PositionRegion}: gripper inside a box above the sink).
\item \pressbutton: press a button while the gripper is tethered to a fixed point by a short rope (\textsc{PositionRegion}: gripper inside a sphere around the rope anchor).
\end{itemize}
The VLM generates these constraints at inference, and \sectapp{app:constraint_eval} audits their accuracy.

\input{figs/contact_links}
\myparagraph{Baseline.}
KALM~\citep{fang2025kalm} is a keypoint-conditioned diffusion-model method that regresses an $SE(3)$ end-effector trajectory from semantic scene keypoints, originally designed for prehensile manipulation.
To evaluate KALM on whole-arm contact, we train it to regress the $SE(3)$ pose of one contact link, chosen per task by the teleoperator, instead of the gripper pose; KALM's distillation, architecture, and training objective are unchanged.
We realize the predicted link-pose trajectory on the robot through the same downstream trajectory optimizer used by \model, with one modification: because the contact target is a 6-DoF link pose rather than a position-and-normal pair, the surface-anchor search is disabled, while constraint handling and beam search are kept identical.
This isolates the comparison to the action representation, link-pose trajectory versus surface-contact trajectory, together with the demonstrations each representation requires, rather than to downstream planning.
The whole-arm systems of \sect{sec:related_work} are not suitable baselines here: they need an object model~\citep{pang2023global,xue2025monte,leve2024explicit} or a simulator and a tactile skin~\citep{barreiros2025learning}, or learn joint-space policies that take no end-effector constraint~\citep{fang2024airexo}.
We compare against KALM because \model builds on its keypoint pipeline, so the two differ only in what they regress from the same keypoints; controls that replace the learned trajectory are in \sect{sec:exp_ablation}.

\myparagraph{Demonstrations.}
For \model, we collect $10$ bare-hand human demonstrations per task using the multi-camera capture pipeline of \sect{sec:traj_gen_pipeline}.
For the baseline, we instead collect $10$ teleoperated demonstrations per task using a GELLO~\citep{wu2024gello} joint-space leader-follower controller, which enables the teleoperator to collect whole-arm demonstrations.
They span varied contact poses and surface regions on the chosen link, so the baseline sees comparable contact diversity.
The two methods thus learn from different demonstration modalities, and we do not separate the two effects: a bare-hand demonstration cannot define the 6-DoF link pose that KALM regresses.

\myparagraph{Protocol and success criterion.}
Each (task, method, setting) cell is evaluated over $10$ rollouts on the physical robot.
Between rollouts we randomize the task-object placement and the robot's home configuration; placements are physically marked and reset so that both methods see the same set, and every placement is kinematically reachable rather than filtered by either planner.
The planner gates every candidate trajectory against the active tolerances $\{\tau_k\}$ (\sect{sec:constraint_gen}), so a rollout either executes a constraint-satisfying trajectory or returns no plan.
We count a rollout as a failure if it returns no plan or does not reach the task subgoal.
The \emph{unconstrained} setting drops the end-effector constraint from the planner, leaving only the contact action; the \emph{constrained} setting enforces it.

\vspacesubsection{Constrained whole-arm manipulation on a 7-DoF xArm}
\label{sec:exp_main}

\tbl{tab:real_world_result} reports the main results: under the unconstrained setting, \model succeeds on most rollouts of every task and is on par with or above KALM throughout, while KALM is less reliable on the harder tasks (\eg, $4/10$ on \opencoffee), consistent with whole-arm contact tolerating small targeting errors once a large contact patch is established.
Under constraints, KALM drops on every task, to half its unconstrained rate on three of four, while \model is essentially unaffected.
The asymmetry is structural: KALM commits to a link-pose trajectory fixed in the demonstrator's embodiment and leaves IK no freedom once a constraint makes it infeasible, whereas \model searches jointly over link, anchor, and joints and reaches the same target through whichever link and anchor satisfy the constraint.
Aggregated over the four tasks, the constrained gap is significant ($p = 4.0\times10^{-5}$) and the unconstrained gap is not ($p = 0.12$); \sectapp{app:results_stats} reports the tests.
The optimizer returned a complete plan on $69/74$ logged planning attempts in the constrained sessions; \sectapp{app:results_failure} tallies the failure modes.
\tbl{tab:replay} shows the same pattern when both planners replay logged scenes offline.

\vspacesubsection{Generalization to new end-effector constraints and embodiments}
\label{sec:exp_multi}

\myparagraph{Multiple end-effector constraints.}
The constraint pipeline of \sect{sec:constraint_gen} is regenerated from the task description at every inference call, so adding a second end-effector constraint is a change of input rather than a change of model.
We stress-test this property on \closefaucet by swapping the pad for a cup that can spill, which layers an \textsc{AxisCone} cup-upright requirement on the \textsc{PositionRegion} above-sink constraint the task already carries.
\model preserves its single-constraint success rate under the tightened two-constraint set, $7/10$ with one constraint and with two (\tbl{tab:ablations}).
Additional constraints enter the optimizer as soft costs paired with hard tolerances, tightening the feasible set without disturbing the contact objective or the policy that produced it.

\myparagraph{Zero-shot cross-embodiment transfer.}
Because the robot's kinematics enter only in the optimizer, the same keypoint policy, reference contact trajectories, and task descriptions run on a 7-DoF Franka FR3 with a Robotiq gripper after changing only the optimizer's robot model (\tbl{tab:real_world_result}, \model (FR3) columns).
The FR3 reaches $75\%$ unconstrained and $68\%$ constrained; the constrained gap to the xArm is concentrated on \opencoffee ($4/10$), where the FR3's thicker links struggle to thread the slim handle, while the other tasks transfer almost unchanged, as analyzed in \sectapp{app:results_cross}.

\vspacesubsection{Ablation study}
\label{sec:exp_ablation}

\input{tabs/2-ablations}
\tbl{tab:ablations} removes one component at a time to show which ones drive \model's constrained performance.

\myparagraph{Sliding anchor.}
To test whether the contact must slide along the link, we pin the surface anchor $\mathbf{u}_t \equiv \mathbf{u}_1$ to its first-waypoint value, which reduces surface-contact IK to fixed-anchor IK.
Success on \opencoffee drops from $7/10$ to $3/10$: the fixed anchor becomes IK-infeasible mid-trajectory under the cup-upright constraint, which a sliding anchor avoids; \sectapp{app:results_ablation} traces the failures.

\myparagraph{Link search.}
To test whether the optimizer must choose the contact link, we pin it to link 5 or to link 6 on \pressbutton.
Link 5 alone gives $6/10$ and link 6 alone $7/10$, against $9/10$ with the search; the two links succeed on largely different placements (two only with link 5, three only with link 6), so each covers placements the other misses, and the search recovers both.

\myparagraph{Geometric primitive.}
To test whether a learned trajectory is needed, we replace the policy with a hand-specified geometric primitive registered to the scene, a line for \pressbutton and an arc for \opencoffee.
The geometric primitive stays close to the policy on the translational press ($7/10$ vs.~$9/10$) but drops from $7/10$ to $4/10$ on \opencoffee, where the capsule carriage translates while the lid rotates less than the handle, so no circular arc fits. The policy instead learns this coupled motion from the demonstrations; \sectapp{app:results_ablation} details the primitive.

\myparagraph{Registered replay.}
To test what the policy adds over the demonstrations themselves, we replay one demonstration registered to the scene by iterative closest point (ICP).
Replay reaches $6/10$ on \opencoffee, so the demonstrated motion does carry the task; the policy reaches $7/10$ without registration or per-task parameters and proposes several candidates per scene instead of one.

%% file: tabs/1-main.tex
\begin{table*}[t]
\centering
\small
\begin{adjustbox}{max width=\textwidth}
\begin{tabular}{@{}lccccccl@{}}
\toprule
\multirow{2}{*}{\textbf{Task}}
& \multicolumn{3}{c}{\textbf{Unconstrained}}
& \multicolumn{3}{c}{\textbf{Constrained}}
& \multirow{2}{*}{\textbf{End-effector constraint}} \\
\cmidrule(lr){2-4} \cmidrule(lr){5-7}
& KALM & \textbf{\model (Ours)} & \model (FR3) & KALM & \textbf{\model (Ours)} & \model (FR3) & \\
\midrule
\opencoffee & 4/10 & 6/10 & 5/10 & 2/10 & 7/10 & 4/10 & Hold cup upright\\
\closecabinet & 7/10 & 7/10 & 8/10 & 5/10 & 8/10 & 7/10 & Hold plate level\\
\closefaucet & 6/10 & 8/10 & 8/10 & 3/10 & 7/10 & 8/10 & Keep gripper above sink\\
\pressbutton & 6/10 & 9/10 & 9/10 & 3/10 & 9/10 & 8/10 & Stay tethered by rope\\
\midrule
\textbf{Aggregate} & 58\% & \textbf{75\%} & 75\% & 33\% & \textbf{78\%} & 68\% & \\
\bottomrule
\end{tabular}
\end{adjustbox}
\caption{\textbf{Real-world results on four whole-arm tasks.}
We evaluate the KALM baseline and our method, \model, on four whole-arm tasks under both unconstrained and constrained settings, each over $10$ rollouts with $10$ demonstrations per task (number of successes shown). KALM and \model (Ours) run on the xArm~7; \model (FR3) runs zero-shot on a Franka FR3 with only the optimizer's robot model changed. 
Under constraints, \model more than doubles KALM's aggregate success ($78\%$ vs.\ $33\%$).}
\label{tab:real_world_result}
\vspace{-1.5em}
\end{table*}

%% file: figs/contact_links.tex
\definecolor{linkFour}{RGB}{234,155,86}
\definecolor{linkFive}{RGB}{90,128,184}
\definecolor{linkSix}{RGB}{140,170,80}
\begin{wrapfigure}{r}{0.30\textwidth}
  \centering
  \includegraphics[width=0.30\textwidth,height=2.4cm,keepaspectratio]{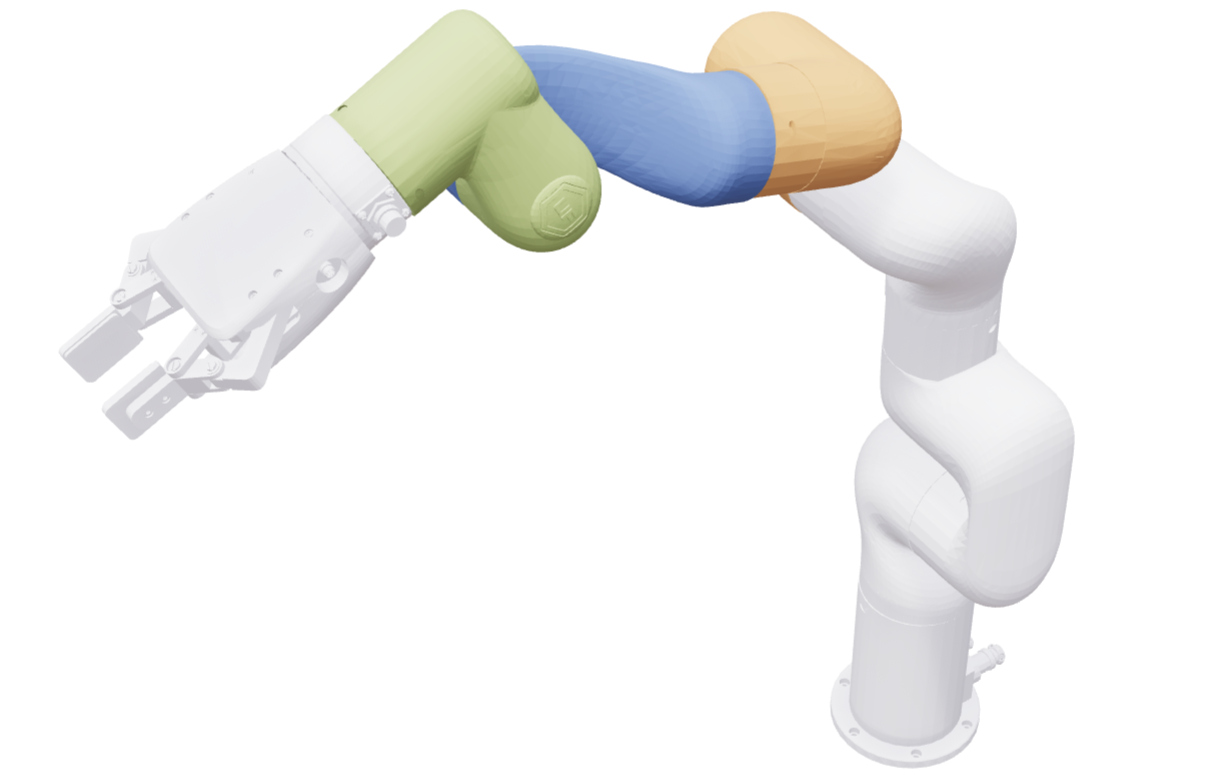}
  \caption{\textbf{Contact links.} Collision meshes of the xArm~7 \textcolor{linkFour}{\textbf{forearm (link 4)}}, \textcolor{linkFive}{\textbf{wrist yaw (link 5)}}, and \textcolor{linkSix}{\textbf{wrist pitch (link 6)}} form the contact surface $\mathcal{M}$.}
  \label{fig:contact_links}
  \vspace{-1.4em}
\end{wrapfigure}

%% file: tabs/2-ablations.tex
\begin{table*}[t]
\centering
\small

\begin{tabular}{@{}lllcc@{}}
\toprule
& & & \multicolumn{2}{c}{\textbf{Success}} \\
\cmidrule(l){4-5}
\textbf{Variant} & \textbf{What changes} & \textbf{Task} & \textbf{Variant} & \textbf{Full \model} \\
\midrule
+ second constraint & cup-upright \textsc{AxisCone} added & \closefaucet & 7/10 & 7/10 \\
\midrule
w/o sliding anchor & anchor fixed at first waypoint & \opencoffee & 3/10 & \textbf{7/10} \\
w/o link search & link fixed to link 5 & \pressbutton & 6/10 & \textbf{9/10} \\
 & link fixed to link 6 & \pressbutton & 7/10 & \textbf{9/10} \\
\midrule
geometric primitive & hand-specified arc as trajectory & \opencoffee & 4/10 & \textbf{7/10} \\
 & hand-specified line as trajectory & \pressbutton & 7/10 & \textbf{9/10} \\
registered replay & ICP-registered demonstration & \opencoffee & 6/10 & \textbf{7/10} \\
\bottomrule
\end{tabular}

\caption{\textbf{Generalization, ablations, and controls.} Each variant runs $10$ rollouts on one constrained task, on the same placements as the full model. A second end-effector constraint leaves success unchanged, while removing an optimizer component or replacing the learned trajectory lowers it.}
\label{tab:ablations}
\vspace{-1.4em}
\end{table*}

%% file: text/5-conclusion.tex
\vspacesection{Conclusion}
\label{sec:conclusion}

We presented \modelfull (\model), a framework that performs a whole-arm contact action without disturbing a primary end-effector task.
Across four real-world tasks on a 7-DoF xArm, \model more than doubles the constrained success rate of a strong link-pose baseline ($78\%$ vs.\ $33\%$), absorbs a second simultaneous constraint without retraining, and transfers zero-shot to a second 7-DoF arm by changing only the optimizer's robot model.
Decoupling what the task requires from how an arm achieves it lets these skills carry over to new constraints and a second arm.

%% file: text/6-limitations.tex
\vspacesection{Limitations}
\label{sec:limitations}
\model plans contact geometrically: it tracks a contact position and normal and models neither force nor object dynamics, so a feasible plan can still slip or under- or over-press, as the failure cases in \sectapp{app:results_failure} show.
That contact is also a single point on a single link; a second simultaneous contact would have its spacing fixed by the arm's link lengths, so the reference trajectory would no longer be robot-agnostic.
The constraints on the other links come from a vocabulary of two primitive types, \textsc{AxisCone} and \textsc{PositionRegion}; a requirement neither expresses cannot be enforced, although each primitive is a short violation function that a VLM could write as code.
Keypoints, trajectory, and constraints are all computed from one RGB-D frame and run open-loop over a short horizon of at most $8$ waypoints, and one inference with a single candidate plan takes $34.6$\,s, most of it in two hosted VLM calls, as the per-stage breakdown in \sectapp{app:results_runtime} shows.
The optimizer's own share is small, but its anchor search grows with the number of contact-capable links; \sectapp{app:limitations} lists further limitations of the evaluation.
Coupling the planner with force or impedance control, enriching the contact representation with multi-link and line- or patch-contact primitives, and closing the loop with tactile or visual replanning remain to be explored.

%% file: text/7-acknowledgments.tex
\section{Acknowledgments}
\label{sec:ack}
The authors thank the USC Center for Advanced Research Computing (CARC) for providing compute resources.
This work was supported by the Office of Naval Research (ONR) Young Investigator Program (YIP) under grant number N00014-26-1-2473.
This research was also supported by the NVIDIA Academic Grant Program using an RTX PRO 6000 Blackwell Max-Q Workstation Edition.

%% file: text/appendix.tex
\appendix

\section*{Appendix}

This supplement collects the details needed to reproduce \model and to verify the claims in the main paper, in the order the main text uses them.
\sectapp{app:method} details the components of \sect{sec:method}: the demonstration data pipeline, the keypoint-conditioned policy, the constraint primitives and their generation, the surface-contact optimizer, and the inference pipeline with its runtime.
\sectapp{app:experiments} describes the real-robot setup and reports the constraint-generation audit, confidence intervals, failure cases, the paired offline replay, cross-embodiment transfer, and the ablation and control details.
\sectapp{app:limitations} lists further limitations.

\section{Method details}
\label{app:method}

\subsection{Demonstration data pipeline}
\label{app:data}

\myparagraph{Multi-view capture.}
A human performs each task bare-handed in front of $C = 3$ calibrated RGB-D cameras, the two scene cameras of \sectapp{app:impl_setup} and the arm's wrist camera, held static during capture, extrinsically calibrated to a common world frame.
We record synchronized RGB-D streams from every camera at $10$\,Hz.
We collect $10$ demonstrations per task, deliberately varying the contact location so the learned policy sees a diversity of contact poses.

\myparagraph{Multi-view fusion.}
Before fusing the per-view MANO tracks of \sect{sec:traj_gen_pipeline}, we shift each view's mesh along its optical axis so that its visible vertices match the sensor depth.
We then fuse the contact fingertip vertex across the $C$ views by per-point weighted-midpoint ray triangulation~\citep{hartley1997triangulation}: each view contributes a back-projected ray through its fingertip estimate, and the fused point minimizes the weighted sum of squared distances to these rays.
Each view's weight is the silhouette IoU of its MANO mesh reprojected into every view against that view's SAM3 hand mask, summed over views, so views whose hand estimate is geometrically consistent dominate the fusion.

\subsection{Keypoint-conditioned policy}
\label{app:impl_policy}

The KALM~\citep{fang2025kalm} policy of \sect{sec:traj_gen_policy} is trained unchanged in architecture on our $(K_\text{obs}, \mathcal{T}^{*})$ pairs with $n_\text{kp}=8$ keypoints, embedding dimension $256$, trajectory length $16$, $100$ diffusion steps, $100$k training iterations, batch size $512$, and AdamW with learning rate $10^{-4}$ and weight decay $5\times10^{-4}$; the KALM baseline uses the same settings.
When several camera views are available, we keep the trajectory from the view whose keypoints have the highest mean feature similarity to the reference.

\subsection{Constraint primitives}
\label{app:impl_constraint}

\sect{sec:constraint_gen} defines the violation $c_k$ and the tolerance $\tau_k$ that every primitive exposes.
Below we specify the two primitives we use, whose per-task parameters \tbl{tab:task_constraints} lists, and the two implementation choices that definition leaves open: the penalty each violation computes and how its tolerance is enforced.
We add each violation as a soft cost to the trajectory optimizer as a native cuRobo constraint, scaled by the quantity it penalizes, radians for \textsc{AxisCone} and meters for \textsc{PositionRegion}, so the two weights are tuned independently.
\input{tabs/app_tasks}
We enforce the tolerance with a \emph{headroom} gate that maps every constraint onto a common normalized scale.
At each waypoint it measures the fraction of the primitive's budget consumed: $\gamma(q)/\alpha$ for \textsc{AxisCone}, or the fraction of the region extent used for \textsc{PositionRegion}.
It discards the trajectory as soon as any waypoint exceeds a cutoff; all reported experiments use a cutoff of $1.0$, \ie, $\tau_k = 0$, so the stated bound is the safety specification itself.
It also defines the ranking's feasibility margin, the smallest budget left over constraints and waypoints.

\myparagraph{\textsc{AxisCone}.}
With parameters $\boldsymbol{\phi}_k = (\ell_k, \mathbf{a}, \mathbf{a}_w, \alpha)$, a constrained link $\ell_k$, a link-local axis $\mathbf{a}$, a target world axis $\mathbf{a}_w$, and a half-angle $\alpha$, the primitive penalizes the angle by which the world-frame axis $R_{\ell_k}(q)\,\mathbf{a}$ of \eqn{eq:fk} leaves the cone around $\mathbf{a}_w$,
\begin{equation}
c_k(q) = \big[\gamma(q) - \alpha\big]_+ , \qquad \gamma(q) = \arccos\big\langle R_{\ell_k}(q)\,\mathbf{a},\,\mathbf{a}_w\big\rangle .
\label{eq:axiscone}
\end{equation}
The penalty is exactly zero inside the cone and grows linearly in the violation angle outside it.
We penalize the angle $\gamma$ directly rather than its cosine.
A cosine-domain hinge has a restoring force proportional to $\sin\gamma$, which is small at the boundary of a tight cone, so small violations are corrected only weakly.
The angle-space L1 penalty instead keeps a constant restoring force per radian, matching the angular quantity the headroom gate checks.

\myparagraph{\textsc{PositionRegion}.}
With parameters $\boldsymbol{\phi}_k = (\ell_k, \mathcal{R})$, a constrained link $\ell_k$ and a convex region $\mathcal{R}$, a sphere or an axis-aligned box in our tasks, the primitive penalizes the exterior distance of the link position $t_{\ell_k}(q)$ of \eqn{eq:fk} from $\mathcal{R}$,
\begin{equation}
c_k(q) = \operatorname{dist}\!\big(t_{\ell_k}(q),\,\mathcal{R}\big) ,
\label{eq:posregion}
\end{equation}
which is zero inside $\mathcal{R}$ and grows linearly with the distance outside.
We evaluate the exterior distance with a pseudo-Huber softened norm, $\sqrt{\lVert\cdot\rVert^2+\epsilon}-\sqrt{\epsilon}$ with $\epsilon=10^{-8}$, so the gradient stays finite at the region boundary.

\subsection{Safety-aware constraint generation}
\label{app:constraint}

\myparagraph{Two-call pipeline.}
The first of the two GPT-5.5 calls of \sect{sec:constraint_gen}, shown in \fig{fig:prompt_scene}, emits for each physical entity it names an ensemble of short SAM3 text prompts ordered from a bare noun to a context-rich phrase.
SAM3 segments each entity in every calibrated view, and the back-projected point clouds are associated across views before the oriented bounding box is fit.
The second call, shown in \fig{fig:prompt_constraint}, follows generic rules that derive the constrained axis from the grasp geometry, keep the constraint set minimal, and add a margin to every region, and it covers latent risks such as spilling, dripping, shedding, contamination, fragility, or loss of task utility.

\myparagraph{Vocabulary.}
We fix the primitive vocabulary to \textsc{AxisCone} and \textsc{PositionRegion} for all reported experiments, but the constraint-instantiation prompt accepts new primitive types without changing any learned component, so richer VLM-driven constraint generation, broader vocabularies, and tighter scene grounding are a natural direction for further exploration.
\input{figs/app_prompts}

\subsection{Surface-contact trajectory optimizer}
\label{app:impl_opt}

We build \model on top of cuRobo~\citep{sundaralingam2023curobo}, whose GPU-batched inverse-kinematics and trajectory-optimization kernels we use to evaluate the full per-link, per-anchor candidate fan-out in parallel.
The surface-contact cost of \eqn{eq:contact_cost}, the constraint primitives of \sectapp{app:impl_constraint}, and the beam-search assembly are our additions on top of this solver.
Below we describe the components whose behavior is not obvious from the main text; all tasks share the cost weights $w_\text{pos} = 10^4$, $w_\text{align} = 2\times10^3$, $w_{\Delta q} = 1$, $w_\text{slide} = 10\,\text{m}^{-1}$, and $w_\text{manip} = 2\times10^{-3}$.

\myparagraph{Reachable anchor set.}
The anchor pruning of \sect{sec:traj_opt} runs before any IK; an anchor counts as occluded when an adjacent link blocks its outward normal within $0.05$\,m, and as buried when it lies inside a neighboring link's signed-distance volume with margin $-0.002$\,m.
At each sliding step we re-sample anchors only within a small Euclidean radius ($0.03$\,m) of the parent anchor, so the search refines the contact locally rather than re-scanning the entire link surface every step.

\myparagraph{Collision buffers.}
Whole-arm contact requires the active link to make contact with the environment, which a standard collision check forbids.
We resolve this by shrinking the contact link's collision spheres with a negative buffer during the sliding stage, admitting controlled surface penetration, while every other link keeps a small positive clearance.
During the free-space approach and the pre-contact pose we use normal positive buffers, keeping the standard collision-avoidance behavior.
The optimizer plans the approach with cuRobo's gradient-based trajectory optimization~\citep{sundaralingam2023curobo}, which can stall in a local minimum near obstacles; when it fails, MoveIt~\citep{coleman2014reducing} plans the approach under the same path constraints with the sampling-based RRTConnect planner, which searches a graph of collision-free configurations instead of descending a cost.

\myparagraph{Surface-contact IK.}
Each (waypoint, anchor) pair is solved with the cost $J_\text{contact}$ of \eqn{eq:contact_cost} from $16$ random IK seeds.
We use a hinge loss on the normal alignment rather than a full orientation cost because surface contact specifies only a position and a normal: this pins five degrees of freedom and leaves the link's roll about the contact normal free as redundancy for the optimizer to spend on the path constraints rather than on an orientation the task never required.
Its deadzone $\theta_\text{dead}$ is the one task-dependent cost parameter: we tighten it for tasks that need a precise contact normal (\eg, pressing a small button) and loosen it for tasks tolerant of contact-patch orientation (\eg, closing the cabinet), trading normal precision for a larger feasible set.
A solution is accepted only if its position residual is within tolerance and it passes the active path-constraint gate of \sectapp{app:impl_constraint}.

\myparagraph{Beam search.}
The beam search of \sect{sec:traj_opt} has width $B=20$ and expands each surviving beam by $32$ candidate anchors per step under the step cost of \eqn{eq:step_cost}; the first waypoint scores $640$ anchors with $4$ pre-contact samples each, and each later step at most $640$ anchors.
The step cost weighs slide far more heavily than joint motion, so the optimizer keeps the anchor stationary by default and spends slide only when the constraint set leaves no stationary-anchor extension feasible, which is the mechanism the ablation in \sect{sec:exp_ablation} isolates.
We additionally cap the per-step joint displacement $\|q_{t+1}-q_t\|$ with a maximum-$\Delta q$ threshold and discard any extension that exceeds it, so beam survivors correspond to continuous, executable arm motion rather than joint-space jumps that happen to score well but would require the wrist to flip between adjacent waypoints.
We downsample the reference contact trajectory $\mathcal{T}^{*}$ before search according to the complexity of the task: $H=4$ for \closecabinet and \pressbutton, and $H=8$ for \opencoffee and \closefaucet.

\myparagraph{Why not a continuous optimization.}
Before the sample-and-search formulation we implemented two continuous alternatives: gradient descent over the joint trajectory with a fixed per-link contact point, and a nonlinear program over joints, per-triangle barycentric coordinates, and softmax blend weights that let the contact point move across a small patch of mesh triangles, solved with IPOPT~\citep{wachter2006implementation}.
The blended contact point and normal are not on the mesh unless the blend is peaky, so intermediate iterates are non-physical and an entropy penalty is needed to drive them back; the discreteness of the mesh and of the link choice re-enters as an outer loop over triangle patches and links, which in practice meant hundreds of multi-start solves per trajectory, most of them far from any feasible contact.
The program also depended on batched per-waypoint IK for its initialization, which is the component the current optimizer keeps; replacing the continuous stage by beam search over that IK output was the change that made planning reliable.
The price is greedy waypoint-by-waypoint assembly with smoothness enforced only by the step cost and per-step joint-displacement cap.

\subsection{Inference pipeline}
\label{app:pipeline}

\alg{alg:inference} summarizes the end-to-end inference described in \sect{sec:inference}.
Its first four lines are the robot-agnostic stages; the robot's kinematics enter only in the per-link search that follows.
\begin{algorithm}[H]
\caption{\textbf{Inference pipeline.} \model inference on a single observation $O=(I, d_\text{task})$: three robot-agnostic stages feed the per-link surface-contact search.}
\label{alg:inference}
\begin{algorithmic}[1]
\Require trained policy $\pi_\theta$, keypoint matcher $\Phi_\text{KP}$, contact links $\mathcal{L}$, per-link anchor sets $\{A_\ell\}$
\Ensure executable joint trajectory with a free-space approach segment
\State $K_\text{obs} \gets \Phi_\text{KP}(I)$, KALM keypoint matching on the best-scoring view \Comment{\sect{sec:traj_gen_policy}}
\State $\mathcal{T}^{*} \sim \pi_\theta(\cdot \mid K_\text{obs})$ \Comment{sample one reference contact trajectory}
\State $\mathcal{S} \gets \textsc{SceneState}(I, d_\text{task})$ \Comment{VLM call 1 $+$ SAM3 3D grounding}
\State $\{\mathcal{C}_k\} \gets \textsc{Constraints}(d_\text{task}, \mathcal{S}, \text{FK})$ \Comment{VLM call 2; \sect{sec:constraint_gen}}
\For{each contact link $\ell \in \mathcal{L}$ \textbf{in parallel}}
  \State sample pre-contact poses; IK toward $(\mathbf{p}^{*}_1,\mathbf{n}^{*}_1)$ over $\mathbf{u}\in A_\ell$; seed $B$ beams
  \For{waypoint $t=2\ldots H$}
    \State extend each beam by IK at neighboring anchors \Comment{drop collision and constraint violators}
    \State keep the top-$B$ extensions under $J_\text{step}$ \Comment{\sect{sec:traj_opt}}
  \EndFor
\EndFor
\State pool per-link trajectories; rank by feasibility margin, then slide, then path length
\State plan a free-space approach to the top candidate's pre-contact pose under $\{\mathcal{C}_k\}$
\State \textbf{return} approach segment $\,\Vert\,$ contact segment
\end{algorithmic}
\end{algorithm}

\subsection{Runtime breakdown}
\label{app:results_runtime}

\fig{fig:runtime} reports the per-stage runtime of a representative inference with a single candidate plan.
The critical path runs through constraint generation, whose three stages, scene decomposition ($10.88$\,s), SAM3 segmentation with 3D grounding ($2.76$\,s), and constraint instantiation ($14.68$\,s), set the optimizer's barrier at $28.31$\,s, while the keypoint policy ($3.73$\,s) and anchor preparation ($6.32$\,s) finish earlier and idle.
The optimizer's stages after the barrier add only $6.32$\,s for a total of $34.63$\,s, so the two hosted-VLM round-trips dominate the wall-clock while the GPU-batched planner is comparatively cheap.
When a task's end-effector constraints are known in advance, as for a task that has already been deployed, the two VLM calls can be skipped; constraints whose parameters depend on the scene, such as the region above the sink or around the rope anchor, still need the SAM3 grounding step to place them. Replanning during execution could likewise reuse the generated constraints and warm-start the optimizer from the previous solution. We did not time either mode on the robot.
The executed trajectory is streamed as points $0.2$\,s apart and interpolated into joint commands at $250$\,Hz.
\input{figs/app_runtime_gantt}

\section{Experiments}
\label{app:experiments}

\subsection{Real-robot setup}
\label{app:impl_setup}

Our main experiments run on a 7-DoF UFactory xArm 7 mounted on a $2\times1$\,m Vention workbench, shown in \fig{fig:app_workspace}.
Two calibrated Intel RealSense D435 RGB-D cameras are fixed to the left and right of the workspace and extrinsically calibrated to a common world frame; together with the arm's wrist camera, they record the bare-hand human demonstrations of \sectapp{app:data}.
Planning runs on a single NVIDIA RTX PRO 6000 GPU.
The cross-embodiment experiments of \sectapp{app:results_cross} use a 7-DoF Franka FR3 with a Robotiq gripper on a $1.2\times0.6$\,m workbench with a similar two-camera rig, shown in \fig{fig:app_franka_setup}.
The same computer is used for planning.
\input{figs/app_setup}

\subsection{Constraint-generation audit}
\label{app:constraint_eval}

\input{tabs/app_vlm_accuracy}
We audited the two-call generator offline on saved camera observations from real sessions: $20$ seeds for each of the four tasks and for the two-constraint variant of \sect{sec:exp_multi}, $100$ generations in total, one GPT-5.5 snapshot, no cache, and no manual filtering.
Each output is scored against a reference constraint for its task: primitive types, constrained link, parameters (axis within $10\degree$, cone angle within a factor of two, region containing $95\%$ of a logged successful trajectory), acceptance of that logged trajectory under forward kinematics, and rejection of adversarial probes that tilt or displace the end-effector.
The deployed generator includes a deterministic feasibility pre-check on the raw specification (correct link, minimum region side, maximum region volume) and re-prompts once on failure; it fired in $8$ of $100$ generations.
\tbl{tab:vlm_accuracy} reports the result: $94$ fully correct, $6$ over-constrained, $0$ looser than deployed.
The six over-constrained outputs misplace the keep-in region: three \pressbutton regions are centered on the parked gripper instead of the rope anchor, and three two-constraint boxes sit above the height at which the arm works; the planner's hard gate rejects them and returns no plan, which is counted as a task failure and is never an unsafe motion.

\subsection{Confidence intervals and tests}
\label{app:results_stats}

\input{figs/app_stats_chart}

\fig{fig:stats} plots the xArm results of \tbl{tab:real_world_result} with Wilson $95\%$ intervals.
Both methods run on the same marked placements, so the primary test is an exact McNemar test on the placement pairs, pooled over the four tasks: under constraints, \model succeeds where KALM fails on $19$ placements and the reverse holds on $1$ ($p = 4.0\times10^{-5}$); without constraints the counts are $11$ and $4$ ($p = 0.12$).
A paired $t$-test over the $40$ placements gives the same conclusion ($p = 7.7\times10^{-6}$ and $0.07$).
Per task, the exact McNemar test in the constrained setting gives $p = 0.125$ (\opencoffee), $0.25$ (\closecabinet), $0.125$ (\closefaucet), and $0.031$ (\pressbutton), without correction for the four tasks, so only the aggregate claim is made.
The sliding-anchor ablation ($7/10$ vs.\ $3/10$) is reported as a mechanism finding, not a significance claim.

\subsection{Failure-case analysis}
\label{app:results_failure}

We report planner failures exactly, from logs, and execution failures qualitatively.

\myparagraph{Planner failures, deployed sessions.}
Mining the logs of the constrained sessions behind \tbl{tab:real_world_result} (four tasks, $74$ planning attempts including re-plans, $148$ candidate contact plans), we find that the optimizer returned a complete plan for $69/74$ attempts ($93\%$) and $124/148$ candidates.
All $24$ failed candidates are contact-stage, and they fall into four causes: $11$ sliding-path collision-gate rejections ($46\%$, all on \closecabinet), $8$ with no collision-free sliding transfer between waypoints ($33\%$), $4$ first-waypoint IK infeasibilities ($17\%$), and $1$ constraint-margin rejection ($4\%$); MoveIt plans any approach the optimizer cannot, as \sectapp{app:impl_opt} describes.

\myparagraph{Execution failures.}
In rollouts that received a feasible plan and still failed, we observed two causes.
\emph{Perception.}
The keypoint matcher mislocalizes the task-relevant geometry, most often on specular or transparent surfaces (the stainless faucet), so the sampled contact trajectory targets the wrong place.
It comes from the depth sensor and the keypoint matcher, either of which can be replaced by a more robust one without changing the rest of the pipeline.
\emph{Open-loop contact.}
The plan is feasible but the contact slips at execution: the arm under- or over-presses because the scene is sensed once and the contact is never corrected from feedback, which is most visible on \opencoffee, where the handle must be lifted through a precise arc.

\subsection{Paired offline planner replay}
\label{app:results_replay}

To extend the sample without robot time, we replayed logged RGB-D observations from the deployed sessions through both planners on the same solver server, with the configuration of the \tbl{tab:real_world_result} sessions restored and verified against each scene's recorded constraint.
Success here is the planner's own flag: every predicted contact waypoint solved by constrained IK within the deployed tolerance, with collision-checked joint paths between waypoints and the scene- and self-collision gates passed.
It measures planning only.
\tbl{tab:replay} reports the results; KALM's rejections are explicit: constrained IK does not converge with the 6-DoF pose target inside the cone, the mechanism claimed in \sect{sec:exp_main}.

\subsection{Cross-embodiment transfer}
\label{app:results_cross}

The Franka FR3 of \tbl{tab:real_world_result} and \sect{sec:exp_multi}, shown in \fig{fig:app_franka_setup}, differs from the xArm in link geometry, contact-capable links, and end-effector frame.
With only the robot model changed, the optimizer selects a different contact strategy than on the xArm on several tasks.

We attribute the constrained gap to the xArm mainly to morphology: the FR3's links are larger and more rounded than the xArm's, so a smaller portion of each link presents a usable contact patch and the surrounding link volume is more prone to colliding with the scene.
This is most apparent on \opencoffee, where the contact must thread a slim machine handle through a precise lifting arc: a thicker, rounder link both struggles to seat on the handle and sweeps a larger collision volume around it.
Line- and patch-contact primitives, which \sect{sec:limitations} lists as future work, could narrow this gap.

\subsection{Ablation and control details}
\label{app:results_ablation}

\myparagraph{Sliding anchor.}
In $5$ of the $7$ failures of the no-sliding variant on \opencoffee the beam dies mid-trajectory: the fixed anchor remains feasible near the start of the contact segment but becomes IK-infeasible as the cup-upright constraint and the reference contact trajectory pull the wrist into incompatible joint configurations, at which point no admissible extension exists and beam search terminates before reaching $t = H$.
Letting the anchor migrate along the link's contact surface trades a small amount of slide for a much larger amount of joint-space slack against the constraint set.

\input{tabs/app_replay}
\myparagraph{Link search.}
The two pinned-link conditions on \pressbutton ran on the same $10$ marked placements as the paired multi-link cell, with a physical reset between conditions.
Four placements were completed by both pinned links, two only by link 5, three only by link 6, and one by neither.
Every pinned plan was feasible and all seven pinned failures occurred at execution, so the per-scene link choice matters for whether the executed contact succeeds, not for whether a plan exists.
The union of the two pinned conditions is an oracle best-of-two; the multi-link result is $9/10$.

\myparagraph{Geometric primitive.}
The primitive replaces the policy by a line (\pressbutton) or circular arc (\opencoffee) whose anchor, axis, sweep, radius, press depth, and offsets are read off one demonstration offline and registered to each scene by ICP; no learned component contributes to the reference trajectory.
This takes six parameters for \pressbutton and seven for \opencoffee, where the policy needs none.
The demonstration also records the contact normal, so the learned trajectory presses in an effective direction, for instance upward under the coffee handle, which a primitive must be told.
On \opencoffee the arc failed by pushing into or slipping off the handle in $5$ of $6$ failures, and by an ICP misregistration in one.

\myparagraph{Registered replay.}
The replay control re-poses the object-frame medoid of the $10$ \opencoffee demonstrations into each scene by ICP and executes it through the unchanged optimizer; registration residuals were $0.71$--$0.81$\,cm and fitness $0.65$--$0.94$.
Following a protocol fixed before the rollouts, the registered trajectory was inspected in the planner preview before each rollout with at most one re-capture allowed on a visibly wrong fit; no re-capture was needed in any of the ten rollouts.
All four failures were at execution (missing or pushing the handle); every replay produced an executable plan.

\section{Additional limitations}
\label{app:limitations}

\myparagraph{Scope of the evaluation.}
Robot-agnosticity is shown on the two fixed-base 7-DoF arms of \sectapp{app:results_cross}; arms with other kinematic structures, mobile bases, and whole-body robots are untested.
Rollouts vary the object placement and the robot's home configuration, not the object instance, so generalization of the keypoints and the contact trajectory to new object instances is not measured.
No experiment pins the roll about the contact normal on its own; \tbl{tab:real_world_result} and the replay of \sectapp{app:results_replay} vary it only jointly with the contact link and the anchor.

\myparagraph{Constraints and parameters.}
``Safety-aware'' refers to the end-effector's own task; the constraints make no guarantee about people near the robot.
The alignment deadzone $\theta_\text{dead}$ and the trajectory horizon $H$ ($4$ or $8$ waypoints) are set per task, as \sectapp{app:impl_opt} lists.

\myparagraph{Demonstration data.}
Bare-hand capture records video of the demonstrator's hands and surroundings; releasing such data requires the demonstrator's consent and removal of identifying content.

%% file: tabs/app_tasks.tex
\begin{table*}[!ht]
\centering
\small
\begin{adjustbox}{max width=\textwidth}

\begin{tabular}{@{}llll@{}}
\toprule
\textbf{Task} & \textbf{Primary end-effector requirement} & \textbf{Primitive} & \textbf{Deployed parameters} \\
\midrule
\opencoffee   & hold cup upright            & \textsc{AxisCone}       & TCP $x$-axis within $25\degree$ of world $+z$ \\
\closecabinet & hold plate level            & \textsc{AxisCone}       & TCP $y$-axis within $25\degree$ of world $+z$ \\
\closefaucet  & keep gripper above sink     & \textsc{PositionRegion} & TCP inside a $0.50\times0.40\times1.40$\,m box over the sink \\
\pressbutton  & stay tethered by rope       & \textsc{PositionRegion} & TCP inside a sphere of radius $0.20$\,m at the anchor \\
\closefaucet, two constraints & above the sink, cup upright & both & $0.50\times0.40\times1.00$\,m box, plus TCP $x$-axis within $25\degree$ of $+z$ \\
\bottomrule
\end{tabular}

\end{adjustbox}
\caption{\textbf{Deployed end-effector constraints.} The primitive and the parameters that ran on the robot for each task; the stated angle or region boundary is the primitive's parameter, enforced as a hard limit ($\tau_k = 0$); TCP is the tool center point.}
\label{tab:task_constraints}
\end{table*}

%% file: figs/app_prompts.tex
\begin{figure}[tp]
\begin{promptlisting}{Prompt 1: Scene decomposition \hfill VLM call 1, input: RGB frame and text}
You are a visual grounding assistant. Your task is to decompose a task description into discrete physical entities and generate a diverse ensemble of short text prompts for each one. These prompts will be used to segment targets via SAM3, which performs best with simple, concrete noun phrases and degrades on overly long or abstract descriptions.

You receive a single RGB image -- the CURRENT FRAME of a robot's workspace.

|Strict usage rules:|
  ~ CURRENT FRAME is your ONLY source of truth for the object's true colour, texture, shape, and identity.
  ~ The robot arm may be visible in the image -- do NOT include it as a target; segment only task-relevant scene objects.

You must follow this three-step process and encode each step explicitly in your JSON output:

|STEP 1 -- Scene Decomposition.|
Read the task description and identify every distinct physical object that must be independently located in the image. Ask yourself: "How many separate things do I need to understand the spatial layout of?" For example:
  ~ "Wiping a countertop" -> countertop, sponge
  ~ "Pushing a box onto a shelf" -> box, shelf
  ~ "Sweeping debris into a dustpan" -> dustpan, debris
Include an object or surface that merely *bounds* where the gripper must remain (a catchment, support, or safe workspace) ONLY when the task explicitly requires bounding the gripper's position. Do not add a speculative "safe region" for every state-sensitive held item; listing an extra surface invites an unnecessary constraint downstream.
Write one sentence naming every object you found. This becomes "scene_decomposition".

Finally, name which single object is the CONTACT TARGET: the scene element the robot must physically act on to accomplish the task. It must be one of the objects you listed. This becomes "contact_target".

|STEP 2 -- Target Count.|
State the exact integer number of distinct objects from Step 1. This becomes "target_count". The "targets" array MUST contain exactly this many entries.

|STEP 3 -- Generate Targets.|
For each object, produce 2 to 5 SAM3 prompts ordered SIMPLEST to MOST COMPLEX:
  1. Bare category name (e.g. "box", "shelf").
  2. Generic near-synonym or shape descriptor (e.g. "wooden crate", "wall rack").
  3. One key distinguishing attribute -- color, material, or spatial relation (e.g. "blue box", "metal shelf").
  4. Context-rich phrase only if truly needed (e.g. "blue box on the right").

Before outputting your answer, assess the scene density. Avoid generating overly broad prompts that would produce over 10 masks (e.g., "box" or "furniture" in a cluttered scene) unless absolutely necessary.

|## Task Description|
@{task_desc}@

|Respond ONLY with valid JSON:|
{
  "scene_decomposition": "<one sentence listing every distinct physical object>",
  "contact_target": "<name of the object the robot must act on>",
  "target_count": <integer>,
  "targets": [
    {
      "object": "<object name>",
      "prompts": ["<prompt1 -- simplest>", "<prompt2>", "..."]
    }
  ]
}
\end{promptlisting}
\caption{\textbf{Scene-decomposition prompt.} 
First VLM call of \sectapp{app:constraint}. 
The model maps the task description to task-relevant entities, adding a bounding surface only when the task requires one, and a SAM3 prompt ensemble per entity, and names the contact target. Task-specific examples are omitted; braces mark variables.}
\label{fig:prompt_scene}
\end{figure}

\begin{figure}[tp]
\begin{promptlisting}{Prompt 2: Constraint instantiation \hfill VLM call 2, input: text only}
Generate safety-aware path constraints for whole-arm manipulation.

The gripper/TCP has a primary task that must remain true at every waypoint while another arm link performs the contact action. Generate constraints only for preserving that primary TCP task, not for contact goals or collision avoidance.

|Primitive choices:|
- axis_cone: preserve orientation for upright/level/spillable/sliding/balanced held objects.
- position_region: preserve position when the held object/material or primary task physically implies a safe workspace, catchment, support, bounded region, or half-space. Open-vocabulary risks include dripping, leaking, shedding, contamination, heat, sharpness, fragility, tipping, dropping, tangling, or loss of task utility.
- Emit the MINIMAL set. Default to exactly ONE constraint. Add a second only if the primary task genuinely cannot be preserved by a single primitive. Do NOT add a position_region if an axis_cone already preserves the held object's state, and do not add a "safe workspace" or "catchment" region that the task does not explicitly require.
- output [] if no TCP path constraint is clearly needed.

|Rules:|
- Constrain only link_tcp.
- Choosing link_axis for an axis_cone (link_axis is a selector: [1,0,0], [0,1,0], or [0,0,1]):
  Reason from the GRASP GEOMETRY of the held object, not from the numbers in the TCP state. The gripper's fingers close along the local y-axis; the palm faces along the local z-axis (approach); the local x-axis is lateral across the palm. Ask: given how an object of this kind is normally held by such a gripper, which LOCAL TCP axis must stay vertical for the object to stay level / not spill / not slide off?
  The TCP state below is ADVISORY ONLY: it may be a PRE-TASK / home configuration recorded before the object was picked up, so its orientation need not be representative of how the object is carried during the task. Use it as a cross-check; if the grasp-geometry reasoning disagrees with the current pose, trust the grasp-geometry reasoning.
- Use scene object OBBs to ground numeric regions, but do NOT copy an OBB extent literally: OBBs are noisy perception output and the TCP must be able to move. Expand the region by 0.10-0.15 m per side beyond the grounding object, and make sure every side of the region is at least 0.25 m so the region can contain an approach-plus-contact motion.
- Keep regions scoped to the task: the robot's task workspace is roughly 1.0 x 0.8 x 0.8 m, and a keep-in region should occupy a modest part of it (typically well under 0.3 m^3). A region that spans the whole visible workspace provides no protection.
- For an "above" constraint, use a box whose x-y footprint covers the relevant support/catchment/safe workspace (plus margin) and whose z range spans the allowed height interval; the z range must start low enough to include the height at which the arm actually works, not only the space above the objects.
- A scene object listed as "NOT LOCALIZED" has no 3D estimate. Never ground a region on a NOT LOCALIZED object: ground it on the nearest localized related object, or fall back to a conservative sphere around the region the TCP must stay in.
- Round numeric values to 2 decimals.
- Output ONLY a YAML list. No prose, no markdown fence.

|Current TCP State (advisory, may be a pre-task pose):|
@{tcp_state}@

|Task:|
@{task_desc}@

|Scene Objects:|
@{scene_objects}@

|YAML schema examples:|
- type: axis_cone
  link: link_tcp
  link_axis: <selector>
  world_axis: <unit vector>
  angle_deg: <degrees>

- type: position_region
  link: link_tcp
  region:
    type: box
    center: [<x>, <y>, <z>]
    quaternion_wxyz: [1.00, 0.00, 0.00, 0.00]
    size: [<dx>, <dy>, <dz>]
\end{promptlisting}
\caption{\textbf{Constraint-instantiation prompt.} 
Second VLM call of \sectapp{app:constraint}. %
Given the task, the tool-center-point (TCP) state from forward kinematics, and the scene oriented bounding boxes, the model derives the constrained axis from the grasp geometry, grounds regions with a margin, and emits the minimal set of typed constraint primitives as a YAML list. Task-specific examples are omitted; braces mark variables.}
\label{fig:prompt_constraint}
\end{figure}

%% file: figs/app_runtime_gantt.tex
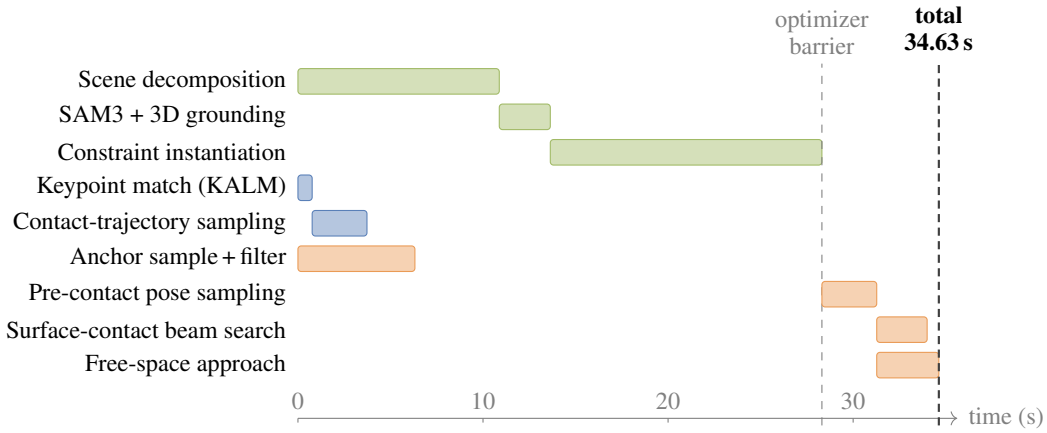
\begin{figure}[H]
\centering
\definecolor{trVL}{RGB}{161,186,102}   %
\definecolor{trPol}{RGB}{ 90,128,184}  %
\definecolor{trOpt}{RGB}{234,155,86}
\def\rtSceneVLM{10.879} %
\def\rtSAM{2.757}       %
\def\rtCons{14.676}     %
\def\rtKp{0.7681}      %
\def\rtTraj{2.9618}    %
\def\rtESDF{1.0013}    %
\def\rtAnchor{6.3200}  %
\def\rtPre{2.9621}     %
\def\rtBeam{2.7239}    %
\def\rtAppr{3.3564}    %
\pgfmathsetmacro{\rtSync}{\rtSceneVLM+\rtSAM+\rtCons}
\pgfmathsetmacro{\tPar}{\rtSync+\rtPre}
\pgfmathsetmacro{\tend}{\tPar+max(\rtBeam,\rtAppr)}
\resizebox{\linewidth}{!}{%
\begin{tikzpicture}[
    x=0.24cm, y=0.46cm,
    lbl/.style={anchor=east, font=\small},
  ]
  \newcommand{\stage}[5]{%
    \node[lbl] at (-0.12,#1) {#5};
    \draw[fill=#4!45, draw=#4!100, rounded corners=1.2pt]
      (#2,{#1-0.36}) rectangle (#3,{#1+0.36});
  }

  \draw[->, gray] (0,0.5) -- (\tend+1.0,0.5);
  \foreach \t in {0,10,20,30}{
    \draw[gray] (\t,0.38) -- (\t,0.5) node[above, font=\footnotesize, gray] {\t};
  }
  \node[font=\footnotesize, gray, anchor=west] at (\tend+1.1,0.5) {time (s)};

  \stage{10}{0}{\rtSceneVLM}{trVL}{Scene decomposition}
  \stage{9}{\rtSceneVLM}{\rtSceneVLM+\rtSAM}{trVL}{SAM3 + 3D grounding}
  \stage{8}{\rtSceneVLM+\rtSAM}{\rtSceneVLM+\rtSAM+\rtCons}{trVL}{Constraint instantiation}
  \stage{7}{0}{\rtKp}{trPol}{Keypoint match (KALM)}
  \stage{6}{\rtKp}{\rtKp+\rtTraj}{trPol}{Contact-trajectory\ sampling}
  \stage{5}{0}{\rtAnchor}{trOpt}{Anchor sample\,+\,filter}
  \draw[dashed, gray] (\rtSync,0.3) -- (\rtSync,10.5)
    node[above, font=\footnotesize, gray, align=center] {optimizer\\barrier};
  \stage{4}{\rtSync}{\tPar}{trOpt}{Pre-contact pose sampling}
  \stage{3}{\tPar}{\tPar+\rtBeam}{trOpt}{Surface-contact beam search}
  \stage{2}{\tPar}{\tPar+\rtAppr}{trOpt}{Free-space approach}
  \draw[densely dashed, black!75, thick] (\tend,0.3) -- (\tend,10.6)
    node[above, font=\footnotesize\bfseries, black, align=center] {total\\34.63\,s};
\end{tikzpicture}%
}
\caption{\textbf{Per-inference runtime breakdown.}
Wall-clock per stage for a representative inference with a single candidate plan; \sectapp{app:results_runtime} gives the numbers.
Shading groups the rows by track: constraint generation (top three rows), the keypoint policy (next two), and the optimizer (bottom four), whose anchor preparation runs concurrently with the other two tracks.
The optimizer's remaining stages start at the dashed barrier once all three tracks finish, with the beam search and the free-space approach overlapped; the right marker is the total.}
\label{fig:runtime}
\end{figure}

%% file: figs/app_setup.tex
\begin{figure}[H]
  \centering
  \begin{subfigure}[b]{0.48\columnwidth}
    \centering
    \includegraphics[height=4.5cm,keepaspectratio]{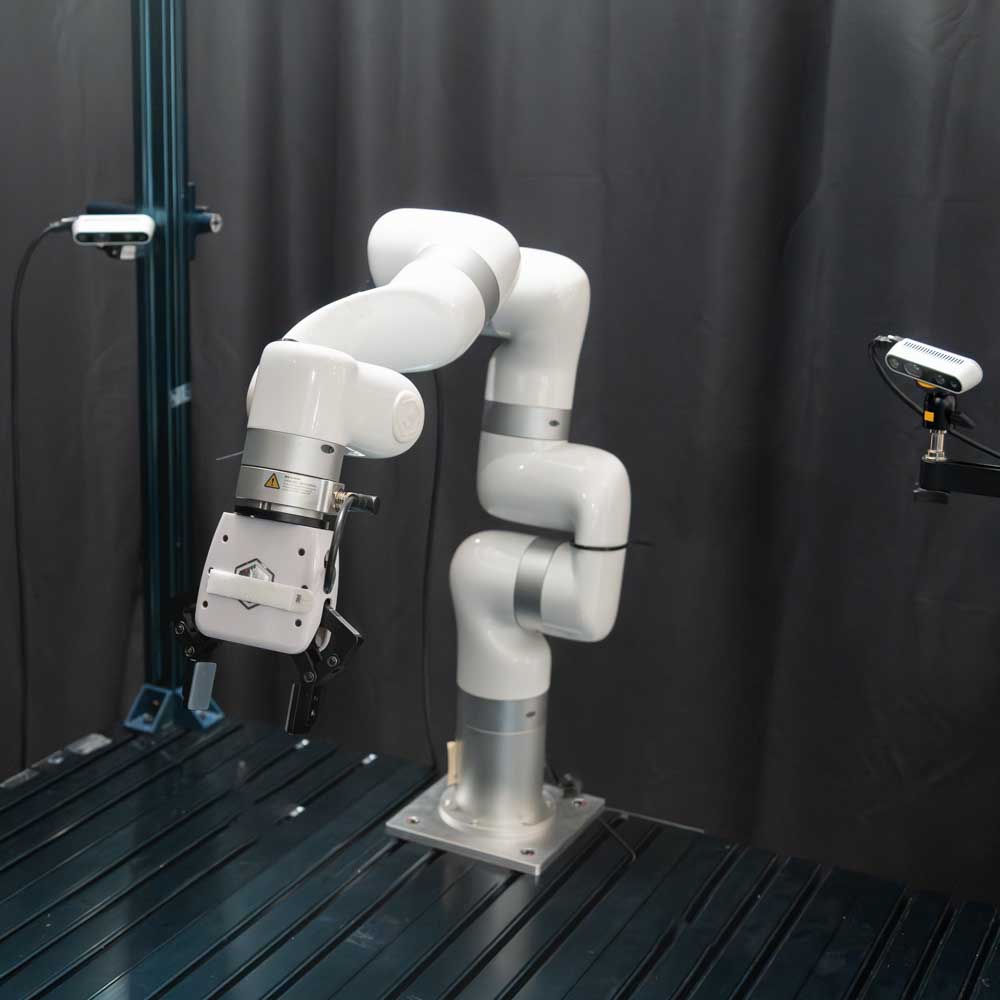}
    \caption{xArm 7 workspace}
    \label{fig:app_workspace}
  \end{subfigure}
  \hfill
  \begin{subfigure}[b]{0.48\columnwidth}
    \centering
    \includegraphics[height=4.5cm,keepaspectratio]{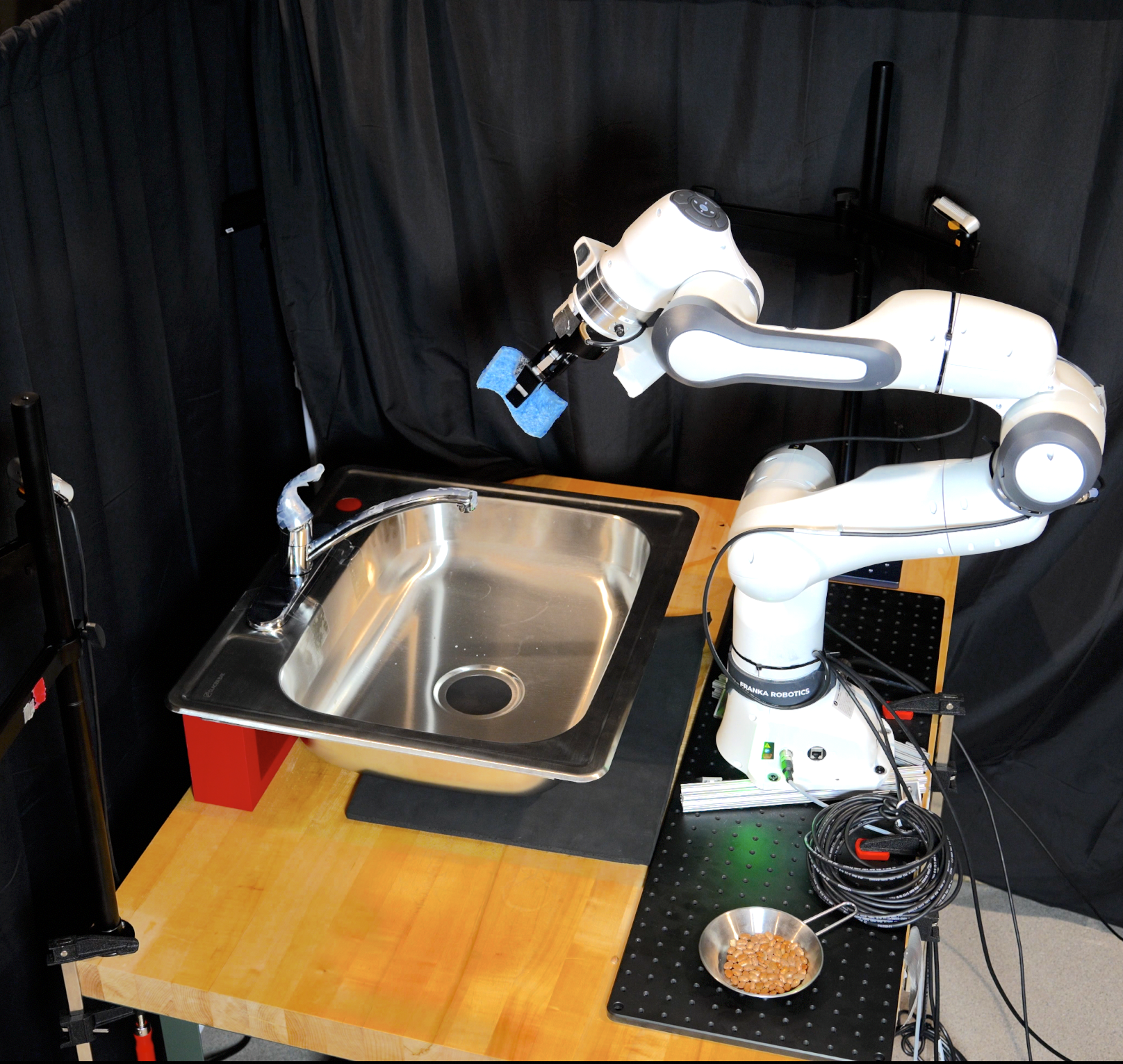}
    \caption{Franka FR3 setup}
    \label{fig:app_franka_setup}
  \end{subfigure}
  \caption{\textbf{Real-robot setups.}
  (\subref{fig:app_workspace})~the 7-DoF xArm 7 used in the main experiments;
  (\subref{fig:app_franka_setup})~the 7-DoF Franka FR3 with a Robotiq gripper used for the cross-embodiment transfer of \sectapp{app:results_cross}.}
  \label{fig:app_setup}
\end{figure}

%% file: tabs/app_vlm_accuracy.tex
\begin{table*}[!ht]
\centering
\small
\begin{adjustbox}{max width=\textwidth}

\begin{tabular}{@{}lcccccc@{}}
\toprule
\textbf{Setting} & \textbf{Parses} & \textbf{Correct type} & \textbf{Fully correct} & \textbf{Over-constrained} & \textbf{Looser than deployed} & \textbf{Re-prompts} \\
\midrule
\opencoffee (\textsc{AxisCone}) & 20/20 & 20/20 & \textbf{20/20} & 0/20 & 0/20 & 0/20 \\
\closecabinet (\textsc{AxisCone}) & 20/20 & 20/20 & \textbf{20/20} & 0/20 & 0/20 & 0/20 \\
\closefaucet (\textsc{PositionRegion}) & 20/20 & 20/20 & \textbf{20/20} & 0/20 & 0/20 & 2/20 \\
\pressbutton (\textsc{PositionRegion}) & 20/20 & 20/20 & \textbf{17/20} & 3/20 & 0/20 & 0/20 \\
\closefaucet, two constraints & 20/20 & 20/20 & \textbf{17/20} & 3/20 & 0/20 & 6/20 \\
\midrule
\textbf{Total} & 100/100 & 100/100 & \textbf{94/100} & 6/100 & 0/100 & 8/100 \\
\bottomrule
\end{tabular}

\end{adjustbox}
\caption{\textbf{Constraint-generation audit.} $100$ unfiltered generations ($20$ seeds per setting) scored against a reference constraint for each task.
``Parses'': the output is a valid primitive list; ``correct type'': the primitive types match the reference; ``fully correct'' requires the right primitive types, the right link, parameters within tolerance, acceptance of a logged successful trajectory, and rejection of adversarial probes; ``over-constrained'': the output rejects that trajectory; ``looser than deployed'': the output accepts it but is not fully correct; ``re-prompts'': the feasibility pre-check fired.
Every incorrect output is over-constrained and is converted by the planner's hard gate into ``no plan''; none is looser than the deployed constraint.}
\label{tab:vlm_accuracy}
\end{table*}

%% file: figs/app_stats_chart.tex
\begin{figure}[H]
\centering
\includegraphics[width=0.52\textwidth]{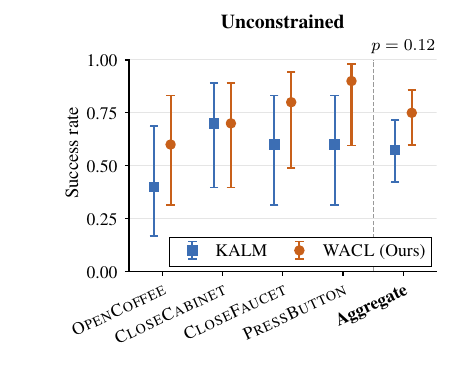}%
\hspace{-0.04\textwidth}%
\includegraphics[width=0.513\textwidth]{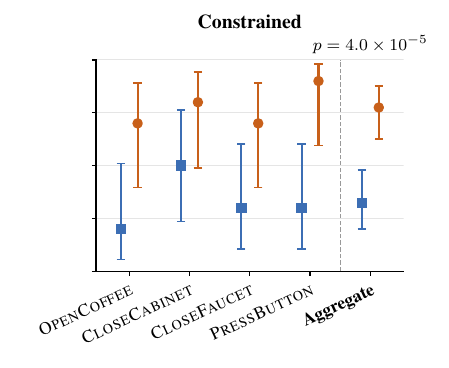}
\caption{\textbf{Success rates with Wilson $95\%$ intervals.} The xArm results of \tbl{tab:real_world_result}, $10$ rollouts per task and $40$ in aggregate. The aggregate constrained intervals are disjoint; the unconstrained ones overlap, consistent with the paired tests in the text; $p$ is the exact McNemar test on the $40$ placement pairs.}
\label{fig:stats}
\end{figure}

%% file: tabs/app_replay.tex
\begin{table*}[!ht]
\centering
\small
\begin{adjustbox}{max width=\textwidth}

\begin{tabular}{@{}lcccccc@{}}
\toprule
& \multicolumn{3}{c}{\textbf{Constraint on}} & \multicolumn{3}{c}{\textbf{Constraint off}} \\
\cmidrule(lr){2-4} \cmidrule(l){5-7}
\textbf{Task} & \textbf{KALM} & \textbf{\model} & \textbf{$p$} & \textbf{KALM} & \textbf{\model} & \textbf{$p$} \\
\midrule
\opencoffee & 2/30 & \textbf{26/30} & $1.2\times10^{-7}$ & 7/30 & \textbf{30/30} & $2.4\times10^{-7}$ \\
\closefaucet & 3/30 & \textbf{30/30} & $1.5\times10^{-8}$ & 29/30 & \textbf{30/30} & $1.0$ \\
\bottomrule
\end{tabular}

\end{adjustbox}
\caption{\textbf{Paired offline planner replay.}
Each row replays the same $30$ logged RGB-D scenes through both planners on the same solver, with the end-effector constraint on and off, and counts complete constraint-satisfying plans; $p$ is the exact McNemar test and nothing is executed.
The \opencoffee scenes are those of \tbl{tab:real_world_result}.
Switching the constraint off helps both planners.
On \closefaucet it closes the gap, while on \opencoffee KALM still fails on most scenes, so its committed 6-DoF pose limits it even without the constraint.}
\label{tab:replay}
\end{table*}